\documentclass[pdflatex,sn-basic]{sn-jnl}% Basic Springer Nature Reference Style/Chemistry Reference Style

\usepackage{graphicx}%
\usepackage{multirow}%
\usepackage{amsmath,amssymb,amsfonts}%
\usepackage{amsthm}%
\usepackage{mathrsfs}%
\usepackage[title]{appendix}%
\usepackage{xcolor}%
\usepackage{textcomp}%
\usepackage{manyfoot}%
\usepackage{booktabs}%
\usepackage{algorithm}%
\usepackage{algorithmicx}%
\usepackage{algpseudocode}%
\usepackage{sourcecodepro}
\usepackage[T1]{fontenc}
\usepackage[utf8]{inputenc}
\usepackage{rotating}
\usepackage{listings}
\usepackage{tabularx}
\usepackage[most]{tcolorbox}
\usepackage[acronym,toc,nopostdot]{glossaries}
\newacronym{ai}{AI}{Artificial Intelligence}
\newacronym{ml}{ML}{Machine Learning}
\newacronym{lp}{LP}{Logic Programming}
\newacronym{ilp}{ILP}{Inductive Logic Programming}
\newacronym{plp}{PLP}{Probabilistic Logic Programming}
\newacronym{hplp}{HPLP}{Hierarchical Probabilistic Logic Program}
\newacronym{pi}{PI}{Predicate Invention}
\newacronym{asp}{ASP}{Answer Set Programming}
\newacronym{llm}{LLM}{Large Language Model}
\newacronym{fm}{FM}{Foundation Model}
\newacronym{tii}{TII}{Technology Innovation Institute}
\newacronym{lff}{LFF}{Learning From Failures}
\newacronym{cot}{CoT}{Chain-of-Thought}
\newacronym{lfit}{LFIT}{Learning From Interpretation Transitions}
\newacronym{nesy}{NeSy}{Neuro-Symbolic}
\newacronym{amr}{AMR}{Abstract Meaning Representation}

\definecolor{black}{rgb}{0.0, 0.0, 0.0}

\newcommand{\review}[1]{\textcolor{black}{#1}}

\newtheorem{casestudy}{Case Study}
\newtheorem{example}{Example}%

\begin{document}

\title[Predicate Renaming via Large Language Models]{Predicate Renaming via Large Language Models\\{\normalsize Preprint}}

%%=============================================================%%
%% GivenName	-> \fnm{Joergen W.}
%% Particle	-> \spfx{van der} -> surname prefix
%% FamilyName	-> \sur{Ploeg}
%% Suffix	-> \sfx{IV}
%% \author*[1,2]{\fnm{Joergen W.} \spfx{van der} \sur{Ploeg} 
%%  \sfx{IV}}\email{iauthor@gmail.com}
%%=============================================================%%
% EG, TR, FR, KI

\author*[1,4]{\fnm{Elisabetta} \sur{Gentili}}\email{elisabetta.gentili1@unife.it}

\author[2,4]{\fnm{Tony} \sur{Ribeiro}}\email{tony.ribeiro@ls2n.fr}

\author[3]{\fnm{Fabrizio} \sur{Riguzzi}}\email{fabrizio.riguzzi@unife.it}

\author[4]{\fnm{Katsumi} \sur{Inoue}}\email{inoue@nii.ac.jp}

\affil*[1]{\orgdiv{Department of Engineering}, \orgname{University of Ferrara}, \orgaddress{\street{Via Saragat 1}, \city{Ferrara}, \postcode{44122}, \country{Italy}}}

% Nantes Université, École Centrale Nantes, CNRS, LS2N, UMR 6004, F-44000 Nantes, France
\affil[2]{\orgname{Nantes Université, École Centrale Nantes, CNRS, LS2N}, \orgaddress{\street{UMR 6004}, \postcode{F-44000} \city{Nantes}, \country{France}}}

\affil[3]{\orgdiv{Department of Mathematics and Computer Science}, \orgname{University of Ferrara}, \orgaddress{\street{Via Machiavelli 30}, \city{Ferrara}, \postcode{44121}, \country{Italy}}}

\affil[4]{\orgname{National Institute of Informatics}, \orgaddress{\street{2-1-2 Hitotsubashi, Chiyoda-ku}, \city{Tokyo}, \postcode{101-8430}, \country{Japan}}}

%%==================================%%
%% Sample for unstructured abstract %%
%%==================================%%

\abstract{
In this paper, we address the problem of giving names to predicates in logic rules using Large Language Models (LLMs). In the context of Inductive Logic Programming, various rule generation methods produce rules containing unnamed predicates, with Predicate Invention being a key example. This hinders the readability, interpretability, and reusability of the logic theory. Leveraging recent advancements in LLMs development, we explore their ability to \review{process} natural language and code to provide semantically meaningful suggestions for giving a name to unnamed predicates. The evaluation of our approach on some hand-crafted logic rules indicates that LLMs hold potential for this task.
}

\keywords{Large Language Models, Inductive Logic Programming, Predicate Invention, Code Refactoring}

%%\pacs[JEL Classification]{D8, H51}

%%\pacs[MSC Classification]{35A01, 65L10, 65L12, 65L20, 65L70}

\maketitle

\section{Introduction}\label{sec:intro}

Logic is an effective approach to modeling domains with complex relationships among entities.
\gls{lp} has been a powerful declarative paradigm based on formal logic since the 70s (\cite{kowalski1988logicprog}). 
Contrary to the traditional imperative programming paradigm, where the focus is on \textit{how} the problem should be solved, in logic programming the program is a description of \textit{what} is the problem.
Logic programs represent knowledge of a domain using facts and rules to state what is true. An \gls{lp} system can derive new knowledge by means of logical inference and answer users' queries about knowledge not explicitly encoded in the program. 
Examples of \gls{lp} languages include Prolog (\cite{colmerauer1996birth}), Datalog (\cite{abiteboul1995foundations}), and Answer Set Programming (\cite{gelfond1988stablemodelsem}).
Since the inference procedure is independent of the knowledge base, logic programs are generally flexible, reusable, and easy to maintain and adapt as new requirements arise. 
New knowledge can be added without changing the inference algorithm.
Despite the many advantages, \gls{lp} also has some challenges.
The knowledge base needs to be carefully structured. 
%Another challenging aspect is when rule-based logic systems output a set of rules among which there are some unnamed predicates. These predicates have been invented or inserted by the systems to adhere to some language bias or to create simpler rules.
To mitigate this problem, \gls{ilp} aims at automatically inducing knowledge bases from data (\cite{muggleton2012ilp}).
Various \gls{ilp} systems have been proposed and have found successful applications (\cite{cropper2022ilp30}) in 
bioinformatics and drug design (\cite{srinivasan1996theories,srinivasan1997carcinogenesis,inoue2013completing}), 
natural language processing (\cite{mooney1996inductive}),
games (\cite{law2014inductive,basu2021hybrid}),
data curation and transformation (\cite{gulwani2011automating}),
explainable \gls{ai} (\cite{zhang2025ilpxai}),
and learning from trajectories (\cite{inoue2014learning}).

Some systems are capable of performing predicate invention, i.e., introducing new unnamed predicates into the learned program to adhere to a language bias or create simpler rules.
Examples of such systems are \textsc{POPPI} (\cite{cropper2021predicate}), \textsc{PHIL} (\cite{nguembang2021learning}), Nesy-$\pi$ (\cite{sha2024neuro}), and many others. 
The presence of such predicates poses a big challenge for both programmers and final users, since it is difficult to understand their intended meaning, especially when the output theory is composed of a large number of rules.
Additionally, manually renaming them is usually unfeasible, since it requires the availability of a domain expert.
In fact, unnamed predicates not only complicate human understanding of the logic program but also hinder the debugging process for programmers.

%In such scenarios, a promising solution in addressing this task could be using \glspl{llm}, given their advanced natural language processing capabilities.
In this paper, we thus propose to use \glspl{llm} to provide names to these predicates.
In the last few years, \glspl{llm} (\cite{alammar2024hands}), a type of generative \gls{ai}, have proven to be highly performing in a wide array of tasks across many domains, including natural language processing, art, and data science, often producing human-like results (\cite{brown2020language,kaddour2023challenges}).
\glspl{llm} are based on deep learning architectures, typically using transformer models (\cite{liu2024understanding, NIPS20173f5ee243}), and are pre-trained on massive amounts of data, enabling them to \review{process} and generate natural language and other types of contents to perform a wide range of tasks, such as text, audio, and video generation, text summarization, code generation, chatbots and query answering. 
Therefore, the goal of this work is to exploit \glspl{llm} to generate semantically meaningful predicate names, which should be naturally understood by the user.
To our knowledge, no systems currently perform automatic renaming of unnamed predicates.

The rest of this paper is structured as follows: 
Section \ref{sec:prelim} provides an overview of the relevant background;  
Section \ref{sec:relwork} presents a review of related work;
Section \ref{sec:method} outlines the methodology used in the study;
Section \ref{sec:exp} describes the experimental setup and approach, and presents the results of the experiments;
Section \ref{sec:disc} discusses the findings and limitations of this work.
Finally, Section \ref{sec:concl} summarizes key findings and suggests directions for future research.

\section{Preliminaries}\label{sec:prelim}

\subsection{Unnamed Predicates}
\gls{ilp} systems may sometimes introduce invented predicates that are not explicitly defined in the knowledge base or problem description.
%These predicates are automatically generated to capture intermediate relationships, simplify complex rules, or make the program more modular and flexible.
There are many reasons to invent new predicates, such as capturing intermediate relationships, simplifying complex rules, enhancing the program's modularity and flexibility, expressing a new concept, improving the accuracy while reducing the complexity of a hypothesis, or when the given vocabulary is insufficient to express the target theory.
However, invented symbols have a placeholder name, such as ``\textit{inv}''. The lack of a meaningful name makes it difficult to understand the intended use of the predicate and its reusability.

In \gls{ilp} (\cite{muggleton1991inductive}), \gls{pi} was first introduced in inverse resolution (\cite{muggleton1988machine}), and early works have been summarized in \cite{stahl1993predicate}.
\gls{pi} refers to those techniques that enable systems to automatically discover new predicates, which are not part of the initial knowledge base (\cite{stahl1995appropriateness}). 
One of the most straightforward examples of \gls{pi}, shown in Example \ref{ex:parent}, is the invention of the predicate \texttt{parent} to learn the definition of \texttt{grandparent}, given only the predicates \texttt{mother} and \texttt{father}: 

\begin{example}\label{ex:parent}
Example of \gls{pi}, where the invented predicate \texttt{inv} represents the parent relation.
\begin{lstlisting}[breaklines=true]
grandparent(A,B) :- inv(A,C),inv(C,B).
inv(A,B) :- mother(A,B).
inv(A,B) :- father(A,B).
\end{lstlisting}
\end{example}

In this way, a more compact theory has been created.
Despite gaining increasing attention in recent developments, \gls{pi} is still not widely spread (\cite{cropper2022ilp30}).
As a matter of fact, the invention of new, useful predicates is not a trivial task (\cite{muggleton2012ilp}), as it is often challenging to determine when and if they are necessary, how many arguments they should have (i.e., their arity), and what their order and type should be (\cite{stahl1995appropriateness,kramer1995predicate}).

\textsc{METAGOL} (\cite{muggleton2015meta}) is a metarule-based approach that seeks to overcome this limitation by using metarules to determine and restrict the syntax of rules, although they should be predefined by the user.

\review{
\textsc{POPPER} (\cite{cropper2021learning}) is an \gls{ilp} system based on the \gls{lff} framework, which decomposes the ILP problem into three steps: generate, test, and constrain. In each cycle, \textsc{POPPER} generates a logic program based on current constraints, evaluates it against examples, and, if the hypothesis is incomplete or inconsistent, derives new constraints from the failure. This iterative refinement continues until a consistent and complete program is found.
\textsc{POPPI} (\cite{cropper2021predicate}) is an \gls{ilp} method that supports the creation of new predicates without needing human intervention to specify the arity or metarules. 
\textsc{POPPI} thus extends \textsc{POPPER} by enabling automatic PI, casting the \gls{ilp} problem as an \gls{asp} task. 
}

\cite{sha2024neuro} proposed NeSy-$\pi$, a \gls{nesy} approach for the classification of complex visual scenes, that processes visual inputs with deep neural networks and invents new concepts.

% In [https://link.springer.com/chapter/10.1007/978-3-540-85928-4_10] the authors used the Hyper ILP system to automatically discover new useful concepts in a robotic domain.

Nevertheless, the discovery of new predicates is not limited to \gls{pi}.

In the context of \gls{plp} (\cite{riguzzi2023foundations}), \glspl{hplp} (\cite{fadja2017deep}) are a restriction of Logic Programs with Annotated Disjunctions (LPADs) (\cite{vennekens2004logic}).
In \glspl{hplp} clauses and predicates are hierarchically organized.
Parameter learning for HIerarchical probabilistic Logic programs (PHIL) (\cite{nguembang2021learning}) is an algorithm for learning the parameters of \glspl{hplp}.  
Given a specific form of LPADs defining the target predicate \textit{r} in terms of input predicates, for which the definition is given as input and is certain, and hidden predicates, defined by the rules of the program, PHIL learns the probabilities of atoms for the target predicate. Each clause in the program has a single atom in the head annotated with a probability.
These programs can be divided into layers. Each layer defines a set of hidden unnamed predicates in terms of either the predicates from the layer below or the input predicates. 
In Example \ref{ex:phil} from \cite{nguembang2021learning} in the UWCSE domain (\cite{kok2005learning}) we can see an example of such predicates.
The first clause $C\_1$ states that the probability of a student (A) being advised by a professor (B) is 0.3 if both A and B have worked on a project (C), and there is a relation $r\_1\_1$ between the student, the professor, and the project. 
The relation $r\_1\_1$ is defined in clause $C\_1\_1\_1$, which states that $r\_1\_1$ holds with a probability of 0.2 if there is a publication P authored by both A and B and produced within project C. $r\_1\_1$ can then be renamed as ``collaboretedOn''.
We can thus say that a student (A) is advised by a professor (B) if there is a project (C) on which both have worked, and A and B have collaborated on project C.

The first clause $C\_2$ states that a student (A) is advised by a professor (B) with a probability of 0.6 if A is a teaching assistant (ta) for course C, which is taught by B. 

\begin{example}\label{ex:phil}
Unnamed predicate r\_1\_1 created by PHIL in the learning process.
\begin{lstlisting}[breaklines=true]
C_1 = advised_by(A,B) : 0.3 :- student(A), professor(B), project(C,A), project(C,B), r_1_1(A,B,C).
C_2 = advised_by(A,B) : 0.6 :- student(A), professor(B), ta(C,A), taughtby(C,B).
C_1_1_1 = r_1_1(A,B,C) : 0.2 :- publication(P,A,C),publication(P,B,C).
\end{lstlisting}
\end{example}

In the context of abductive logic programming (\cite{kakas1992abductive}), the approach proposed by (\cite{sato2018abducing}) aims to abduce relations by leveraging linear algebra in vector spaces. Hidden relationships between entities are learnt through matrix operations, making it useful for applications where explicit rules are hard to define.
Focusing on Datalog programs with binary predicates, the authors address both non-recursive and recursive cases. 
For instance, in the non-recursive case, given two relations $r_1(X,Y)$ and $r_3(X,Z)$, represented as adjacency matrices $\textbf{R}_1$ and $\textbf{R}_3$ respectively, the system learns a new relation $r_2(Y,Z)$, represented by matrix $\textbf{R}_2$, that explains how $r_1$ and $r_3$ interact. This is done by solving the matrix equation $\textbf{R}_3 = min_1(\textbf{R}_1\textbf{R}_2)$, where $min_1(x)$ is a nonlinear function, so that $r_3(X, Z) \Leftrightarrow \exists Y r_1(X,Y) \land r_2(Y,Z)$. Although $r_1$ and $r_3$ are known, $r_2$ is not, therefore it could be useful to find a meaningful name for easier reuse.
For example, the authors have applied this method to a subset of the FB15k (\cite{bordes2013translating}) knowledge graph, involving person and film.
By considering the relations \textit{language(X,Z)} and \textit{genre(X,Y)}, the following rule can be discovered:
$$language(X,Z) \Leftrightarrow genre(X,Y) \land r_{2\_abd}(Y,Z).$$
Here, the abduced relation $r_{2\_abd}$, which isn’t originally in the graph, could be interpreted as \textit{genre\_lang}, implying that genre and language are somewhat related to each other.
However, such an interpretation on the new predicate $r_{2\_abd}$ as well as the predicate name \textit{genre\_lang} have to be given by the user.

\subsection{Overview of Large Language Models}
% background on LLMs
% GenAI -> FMs -> LLMs
Generative \gls{ai} (\cite{feuerriegel2024generative}) is the subset of \gls{ai} techniques that generate new content in the form of text, code, images, audio, and video. 

One type of Generative \gls{ai} are \glspl{fm} (\cite{bommasani2021opportunities}), very large \gls{ml} models trained on enormous quantities of generic and unlabeled data and thus capable of performing a wide range of tasks.

Among \glspl{fm}, \glspl{llm} are based on the transformer architecture (\cite{liu2024understanding,NIPS20173f5ee243,alammar2024hands}). They are composed of multiple layers of neural networks, with an encoder and a decoder (\cite{raffel2020exploring,yang2020decoder}) that process sequences of text, \review{capturing} the connections and the relationships between words. Thanks to the self-attention mechanism (\cite{NIPS20173f5ee243}), the model can consider distant tokens as well.
To represent words, \glspl{llm} use multi-dimensional vectors, called word embeddings, in a way that keeps related words close to each other in the embedding space.

These models often have billions of parameters and can be trained on vast amounts of data, enabling them to learn not only the grammar but also the semantics of a language.
Given an input sequence, parameters are continuously adjusted during the training phase to maximize the likelihood of the next tokens.

\glspl{llm} have gained popularity in recent years (\cite{brown2020language}) thanks to their ability to generate meaningful content from little input.

Various \glspl{llm} have been specifically trained and used for automatic code generation showing good performances (\cite{brown2020language,kaddour2023challenges}), such as OpenAI’s Codex (\cite{chen2021evaluating}), and its GitHub implementation, Copilot\footnote{https://copilot.github.com}.
For example, \cite{pearce2023examining} have tested Codex to determine if it can identify security flaws and fix them, obtaining positive results on hand-crafted datasets; however, real-world scenarios proved to be more challenging.
\cite{moradi2023copilot} demonstrated that Copilot can become a useful tool for software developers for implementing simple procedures, speeding up their work. Given their expertise, even if the generate code contains bugs, they should be able to identify and fix them.

% prompt engineering --> maybe not here
\subsubsection{Prompt Engineering}
Prompting is crucial when working with \glspl{llm} to get the most relevant and useful answers. A carefully crafted prompt can significantly simplify the process by precisely defining the goal we aim to achieve.
A basic prompt usually includes only the instructions for the task at hand, such as ``summarize'' or ``classify''. Other elements that can be included are: context; the audience; the role of the model; input data; output format; and examples of inputs and outputs (\cite{alammar2024hands}). The tone of the response can also be controlled. Additionally, the order also matters: information in the initial or in the final part of the prompt is often given more weight than that in the middle. This behavior is referred to as primacy effect and recency effect, respectively (\cite{alammar2024hands},\cite{liu2024lost}).
These components are not always mandatory, but the more specific the prompt, the better the response will be. Giving an overly general prompt to the \gls{llm} may result in its inability to complete the given task.

One or more examples of the desired output can also be included in the prompt. 
This is referred to as in-context learning or few-shot learning (\cite{min2022rethinking, brown2020language}), in which a few examples of the task at hand and a test example are provided, \review{as shown in Example \ref{ex:few_shot_prompt} taken from~\cite{brown2020language}.}
(\cite{touvron2023llama, alammar2024hands}) proved that implementing these strategies leads to improved performances.

\begin{example}\label{ex:few_shot_prompt}
\review{Few-shot prompting example from \cite{brown2020language}.}
\begin{tcolorbox}[colframe=black, colback=white, boxrule=0.4pt, sharp corners]
\textbf{Prompt:} A "whatpu" is a small, furry animal native to Tanzania. An example of a sentence that uses the word whatpu is:

We were traveling in Africa and we saw these very cute whatpus.

To do a "farduddle" means to jump up and down really fast. An example of a sentence that uses the word farduddle is:
\newline

\textbf{Model answer:} One day when I was playing tag with my little sister, she got really excited and she started doing these crazy farduddles.
\end{tcolorbox}
\end{example}

Another prompting technique is \gls{cot} Prompting, in which the output of one prompt is given as input of the next one, creating in this way a chain of prompts. 
\gls{cot} is at the base of \glspl{llm} reasoning (\cite{wei2022chain,kojima2022advances}), as shown by OpenAI's reasoning models (\cite{openai2025gpto3minicard}). Before generating an answer, these models learn to refine their thinking process, correct their mistakes, and explore different paths, building a chain of thought and leading to a higher-quality final result.

%Creating the optimal prompt is a gradual process, done by experimentation, which plays a key role in crafting the most effective one for each use case.
\review{
One of the main challenges in using LLMs is designing prompts that are both clear and effective in guiding the model toward the desired output. The quality of the prompt has a significant impact on the model’s performance, and even small variations in wording or structure may lead to different results. Crafting an optimal prompt is therefore not straightforward, but is a gradual, iterative process that requires careful experimentation to determine what works best for each specific use case.
Another major challenge is handling the inherent non-determinism of \glspl{llm}, as the same prompt can yield different outputs across multiple runs, making it difficult to ensure consistency and reproducibility.
}

\section{Related work}\label{sec:relwork}
\gls{pi} is considered a critical step, yet it is still not widely spread (\cite{cropper2022ilp30}).
Choosing suitable names for newly created predicates is an even more neglected area in \gls{ilp} and has not received much attention so far, with only a few attempts made in the past.

In \cite{inoue2010discovering} 
%the authors used a meta-level abduction approach to discover unknown relations in a graph representing the knowledge of a certain domain. Here, new predicates are invented in the form of existentially quantified variables, however, their names should be defined based on the specific domain.
\review{
the authors adopted a meta-level abduction approach to identify previously unknown relations within a knowledge graph that represents a specific domain. In this setting, new predicates are invented as existentially quantified variables, which act as placeholders for the missing or hidden knowledge. While the system is capable of structurally generating these new predicates, their semantic meaning (i.e., their names) is not determined automatically and must instead be manually defined or interpreted based on the domain context. This highlights a key limitation: the invented predicates are syntactically valid but lack meaningful, human-readable names without further domain-specific interpretation.
}
In the rule abduction approach proposed in \cite{kobayashi2009hypothesis}, the authors built databases related to the cello playing domain to assist performers in assigning meaningful names to new abduced rules.

As the power of \glspl{llm} continues to grow, we could leverage their capabilities to enable them to reason about rules and generate meaningful names for previously unnamed predicates.
While many have highlighted the impressive performances of \glspl{llm} with zero or few-shot prompting (\cite{brown2020language, kojima2022advances, wei2022chain}), a rising number of recent works have explored their reasoning ability on more complex tasks (\cite{li2024llms}) involving common knowledge and planning (\cite{valmeekam2022large,talmor2018commonsenseqa}), math (\cite{cobbe2021training,kojima2022advances}), and symbolic reasoning (\cite{ishay2023leveraging,kojima2022advances,srivastava2022beyond}).
In particular, \glspl{llm} are already widely employed in software engineering (\cite{brown2020language,hou2024se,jiang2024codegen,kaddour2023challenges}), where they typically generate code with semantically meaningful variable names.

\cite{li2024llms} evaluated state-of-the-art \glspl{llm} on an \gls{ilp} benchmark and found their performance lacking when compared to a smaller neural program induction model.
They evaluated \glspl{llm} on two benchmarks, one of which required the model to induce programs to deduce more complex family relationships from some basic ones.
Although this setup is similar to ours, they focused on assessing the relational reasoning capabilities of \glspl{llm} using standard natural language and truth value prompting. In contrast, our goal was to determine if \glspl{llm} could identify names that are relevant to the context of the example.
\cite{han2024inductive} used GPT-3.5 and GPT-4 for property induction, a traditional inductive reasoning task \review{that involves generalizing a property from a few examples to new cases. In their experiments, participants—including both humans and language models—were presented with a set of objects sharing a property and asked to infer whether new objects shared the same property.}
Although GPT-3.5 had difficulty capturing various aspects of human behavior, GPT-4's performance was mostly close to that of humans. 

\cite{yang2024language} investigated the ability of pretrained language models to infer natural language rules from natural language facts, obtaining promising results.
\cite{wang2023hypothesis} proposed an approach to improve \glspl{llm}’ inductive reasoning ability. They generate hypotheses in natural language, and then translate them into specific programs used for making predictions, breaking the task into two levels of abstraction and outperforming current baselines.
In \cite{tarau2025llm} the authors explored techniques to extract and leverage the knowledge embedded in \glspl{llm} by converting it into different types of logic programs, aiming to improve reasoning capabilities and ensure that \gls{llm} outputs are consistent with their intended purposes.
Additionally, \cite{gandarela2024inductive} aimed to determine whether \glspl{llm} can address ILP tasks, analyzing their capabilities and limitations on logic theory induction. Their results indicate that while larger \glspl{llm} can achieve competitive results compared to state-of-the-art \gls{ilp} systems, long predicate relationship chains are still an obstacle for \glspl{llm} in inductive reasoning tasks.
\cite{cunnington2024role} proposed NeSyGPT, a novel architecture that integrates neural and symbolic reasoning, that fine-tunes a vision-language foundation model to extract symbolic features from raw data, and then learns an expressive answer set program using them. The results show that NeSyGPT outperforms various baselines in accuracy and can be used for more complex \gls{nesy} tasks. They thus demonstrated that \glspl{llm} can reduce the need for manual efforts in automating the interface between neural and symbolic components.
Remaining in the \gls{nesy} field, \cite{de2024neurosymbolic} presented a novel approach to expand the knowledge base by integrating \glspl{llm} with structured semantic representations. It employs a state-of-the-art \gls{llm} to produce a natural language description of an input image, which is then transformed into an \gls{amr} graph. This graph is further extended with logical design patterns and layered semantics derived from linguistic and factual knowledge bases.
\cite{kareem2025using} proposed LLM2LAS, a system that learns commonsense knowledge in the form of \gls{asp} programs, from story-based question and answers in natural language, using only a few examples and combining \glspl{llm}' semantic parsing abilities with ILASP (\cite{law2020ilasp}), a tool for learning \gls{asp} programs. 
%Regarding the use of \glspl{llm} in the generation of logic rules, 
The authors of \cite{ishay2023leveraging} used an \gls{llm} to generate \gls{asp} programs representing logic puzzles, given their description in natural language. Here, the authors state that the \gls{llm} successfully produced an \gls{asp}, despite some minor errors that were corrected manually, confirming the possibility of leveraging \glspl{llm} to support the creation of \gls{asp} programs.
Although some findings suggest that \glspl{llm} still struggle with complex reasoning (\cite{creswell2022selection,li2024llms}), this proves that the field yields some potential and is worth further investigation, especially considering the continuous improvements of \glspl{llm}.

\section{Renaming predicates using LLMs}\label{sec:method}

In this section, we outline our approach for assigning meaningful names to unnamed predicates in a set of logical rules.

\subsection{Predicate Renaming Pipeline}

\begin{figure}[h]
    \centering
    \includegraphics[width=\textwidth]{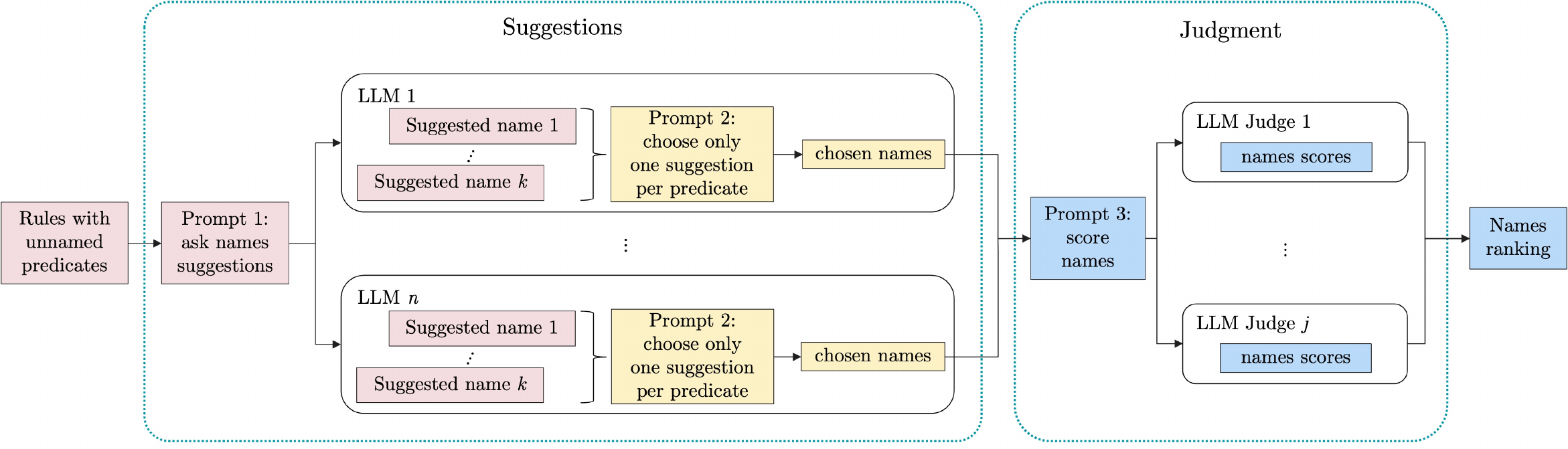} 
    \caption{Diagram illustrating key components and interactions of the approach we propose.}
    \label{fig:idea}
\end{figure}

After deciding the number \textit{n} of \glspl{llm} to query for suggestions,  the number \textit{k} of suggestions they should provide, and  the number \textit{j} of \gls{llm} judges, the pipeline we propose consists of three main steps, as shown in Figure~\ref{fig:idea}:
%Figure~\ref{fig:idea} depicts the pipeline that we propose, which involves three main steps:
\begin{enumerate}
\item Asking the \textit{n} \glspl{llm} to find a meaningful name for each unnamed predicate, for \textit{k} times to tackle possible hallucinations; for these preliminary experiments, we decided to repeat the question three times.
\item Asking the models to choose the most suitable name among their own suggested names.
\item Asking the \textit{j} judges to score the different proposed names, obtaining a ranking.
\end{enumerate}

\subsection{Materials}
To assess the ability of the \glspl{llm} to find correct names for unnamed predicates, we created different examples of logic theories, which represent the possible output of an \gls{ilp} system \review{in a declarative logic programming language}. 
\review{ 
Although constructs that are specific to Prolog were used for these experiments, this approach could be applied to scenarios where other logic programming languages are employed.  
For example, if rules are written in a purely declarative manner, those rules may be commonly used for both Prolog and ASP.
}

In each set of rules, we replaced the true names of some or all predicates with placeholders.
We tested various formats (e.g., A, P, h0, and so on) for placeholders to determine whether the placeholder name influences the suggestions.
Unnamed predicates can occur in the head and/or in the body of one or multiple rules.
\review{For all case studies presented in this paper, the true names of the unnamed predicates are included as comments (e.g., \texttt{\# P = coauthors)} for the reader’s reference. These comments were not included in the prompts provided to the \glspl{llm}.}

\paragraph{Unnamed predicates appearing only in the head}

The first and simplest case study, \textit{coauthors}, consists of a single rule, as shown in Case Study \ref{cs:coauthors}.
Here, \texttt{P} is only one unnamed predicate appearing in the head of a rule, which states that there are two researchers (A and B) and both authored a paper (C); P could thus be renamed as ``coauthors''.

\begin{casestudy}[coauthors] \label{cs:coauthors}
Rules of the coauthors case study.
\begin{lstlisting}[breaklines=true]
P(A,B) :- author(A,C), author(B,C), researcher(A),
researcher(B), paper(C). # P = coauthors
\end{lstlisting}
\end{casestudy}

\paragraph{Unnamed predicates appearing both in the head and in the body}
The second case study, \textit{family}, is described in Case Study \ref{cs:family} and contains several rules representing various family relationships. 
Unnamed predicates appear both in the head and in the body of the rules, with \texttt{h2} being the only exception, as it appears only on the head on a single rule.

\begin{casestudy}[family] \label{cs:family}
Rules of the family case study.
\begin{lstlisting}[breaklines=true]
h0(X,Y) :- mother(X,Y). # h0 = parent
h0(X,Y) :- father(X,Y).
h1(X,Y) :- h0(X,Z), h0(Z,Y). # h1 = grandparent 
ancestor(X,Y) :- h0(X,Y).
ancestor(X,Y) :- h0(X, Z), ancestor(Z,Y).
h2(X, Y, Z) :- ancestor(Z, X), ancestor(Z,Y). # h2 = common_ancestor 
h3(X,Y) :- sister(X,Y). # h3 = siblings 
h3(X,Y) :- sister(Y, X).
h3(X,Y) :- brother(X,Y).
h3(X,Y) :- brother(Y,X).
h4(X,Y) :- h0(PX, X), h0(PY,Y), h3(PX, PY), dif(PX,PY). # h4 = cousins 
\end{lstlisting}
\end{casestudy}

The third case study, \textit{math}, shown in Case Study \ref{cs:math}, is a set of rules representing basic mathematical operations.
Also in this case, unnamed predicates can appear both in the head and in the body of the rules.

\begin{casestudy}[math] \label{cs:math}
Rules of the math case study.
\scriptsize
\begin{lstlisting}[breaklines=true]
A(X) :- integer(X), !. # isNumber
A(X) :- float(X).

P(X,Y) :- A(X), A(Y), X > Y. # greaterThan
Q(X,Y) :- A(X), A(Y), X >=Y. # greaterOrEqual
R(X,Y) :- A(X), A(Y), X < Y. # lessThan
S(X,Y) :- A(X), A(Y), X <=Y. # lessOrEqual
T(X,Y) :- A(X), A(Y), X = Y. # equalTo

B(X) :- A(X), 0 is mod(X,2). # isEven
C(X) :- A(X), not B(X).      # isOdd

D(X,Y) :- ( A(X), R(X,0) ->  # abs
   Y is -X
   ;
   Y is X
).

E(X,Y,Z) :- A(X), A(Y), Z is X + Y. # sum
F(X,Y,Z) :- A(X), A(Y), Z is X - Y. # difference
G(X,Y,Z) :- A(X), A(Y), Z is X * Y. # product
H(X,Y,Z) :- A(X), A(Y), dif(Y,0), Z is X / Y. # quotient

L(X,Y) :- A(X), A(Y), dif(X,0), 0 is mod(Y,X). # isDivisor

M(0,Y,Y):- A(Y), P(Y,0). # gcd
M(X,0,X):- A(X), P(X,0).
M(X,Y,Z):- A(X), A(Y), P(X,0), P(Y,0), R(X,Y), M(Y,X,Z).
M(X,Y,Z):- A(X), A(Y), P(X,0), P(Y,0), P(X,Y), T is X mod Y, M(Y,T,Z).

N(X,0,0). # lcm
N(0,X,0).
N(X,Y,Z) :- 
   A(X), A(Y),
   D(Y,U), D(X,K), 
   M(X,Y,W),
   H(K,W,V), 
   G(U,V,Z).
\end{lstlisting}
\end{casestudy}

\paragraph{Unnamed predicates appearing only in the body}
We also created simpler one-rule examples to investigate the ability of the \glspl{llm} to find a meaningful name for predicates appearing only in the body.
These examples are taken from the previous ones and are shown in Case Study \ref{cs:gp_cs_lcm}: the \textit{grandparent} rule has predicate \texttt{h0} in the body, representing the \texttt{parent} relation; the \textit{cousins} rule with predicate \texttt{h3} representing the \texttt{siblings} relation; and lastly, the \textit{lcm} rule, representing the computation of the least common multiple using the greatest common divisor, uses predicates \texttt{G} and \texttt{H}, being \texttt{product} and \texttt{quotient}, respectively.
In all these cases, the unnamed predicates appear only in the body of the rule.

\begin{casestudy} \label{cs:gp_cs_lcm}
Rules of the \textit{grandparent}, \textit{cousins}, and \textit{lcm} case studies.
\begin{lstlisting}[breaklines=true]
grandparent(X,Y) :- h0(X,Z), h0(Z,Y). # h0 = parent
---------------------------------------------------------
cousins(X,Y) :- parent(PX, X), parent(PY,Y), h3(PX, PY), 
                dif(PX,PY). # h3 = siblings
---------------------------------------------------------
least_common_multiple(X,Y,Z) :- 
   G(X,Y,V), # G = product/multiply
   greatest_common_divisor(X,Y,W),   
   H(V,W,Z). # H = quotient/divide
\end{lstlisting}
\end{casestudy}

\paragraph{Unnamed predicates in a real-world dataset}
In addition, we also tried a real-world knowledge base taken from the mutagenesis dataset (\cite{srinivasan1996theories}), a classic \gls{ilp} benchmark dataset for Quantitative Structure-Activity Relationship (QSAR), that is, predicting the biological activity of chemicals from their physicochemical properties or molecular structure. The goal is to predict the mutagenicity of compounds given their chemical structure. We used the \textit{ring\_theory} provided with the dataset and removed the names of most of the head predicates. The set of rules is extensive, with 24 head predicates to be renamed, such as those in Case Study \ref{cs:muta_example_rules}.

\begin{casestudy}[Mutagenesis] \label{cs:muta_example_rules}
Examples of rules of the Mutagenesis ring theory. Here, the "@>" operator checks if the first term is greater than the second term according to Prolog's standard order of terms.
\begin{lstlisting}[breaklines=true]
anthracene(Drug,[Ring1,Ring2,Ring3]) :-
   benzene(Drug,Ring1),
   benzene(Drug,Ring2),
   Ring1 @> Ring2,
   interjoin(Ring1,Ring2,Join1),
   benzene(Drug,Ring3),
   Ring1 @> Ring3,
   Ring2 @> Ring3,
   interjoin(Ring2,Ring3,Join2),
   \+ interjoin(Join1,Join2,_),
   \+ members_bonded(Drug,Join1,Join2).

no_of_nitros(Drug,No) :-
   setof(Nitro,nitro(Drug,Nitro),List),
   length(List,No).

ring6(Drug,[Atom1|List],[Atom1,Atom2,Atom4,Atom6,Atom5,Atom3],
   [Type1,Type2,Type3,Type4,Type5,Type6]) :-
   bondd(Drug,Atom1,Atom2,Type1),
   memberchk(Atom2,[Atom1|List]),
   bondd(Drug,Atom1,Atom3,Type2),
   memberchk(Atom3,[Atom1|List]),
   Atom3 @> Atom2,
   bondd(Drug,Atom2,Atom4,Type3),
   Atom4 \== Atom1,
   memberchk(Atom4,[Atom1|List]),
   bondd(Drug,Atom3,Atom5,Type4),
   Atom5 \== Atom1,
   memberchk(Atom5,[Atom1|List]),
   bondd(Drug,Atom4,Atom6,Type5),
   Atom6 \== Atom2,
   memberchk(Atom6,[Atom1|List]),
   bondd(Drug,Atom5,Atom6,Type6),
   Atom6 \== Atom3.
\end{lstlisting}
\end{casestudy}

Some of these rules are relatively short (1 to 3 body atoms), while others are much more complex, with more than 5 body atoms.

\review{
\paragraph{Unnamed predicates in an actual \textsc{POPPER} output}
To test our approach on an actual \gls{ilp} theory, we designed a new case study and used \textsc{POPPER} to learn a recursive definition of a reachability predicate over a graph defined by bidirectional (road/2) and unidirectional (one\_way/2) connections. The goal was to induce a definition of reachable/2 using predicate invention, such that \textsc{POPPER} introduces an intermediate invented predicate (e.g., inv1/2) to generalize a single-step path relation.
The resulting program is presented in Case Study~\ref{cs:reachable}.
Given that this case study was specifically designed by the authors for this work, it is not present in the existing literature to the best of our knowledge. As a result, it is unlikely that the LLMs were exposed to it during training.
}

\begin{casestudy}[reachability] \label{cs:reachable}
Rules of the reachability case study.
\begin{lstlisting}[breaklines=true]
inv1(V0,V1):- road(V0,V1).
inv1(V0,V1):- one_way(V0,V1).
reachable(V0,V1):- inv1(V0,V1).
reachable(V0,V1):- reachable(V2,V1),inv1(V0,V2).
\end{lstlisting}
\end{casestudy}

\subsection{Prompts}
% To each \gls{llm} chosen, we ask to find a meaningful name for the predicates with a placeholder. This should be repeated multiple times to tackle potential hallucinations; for these preliminary experiments, we decided to repeat the question three times. After that, we ask to choose the most suitable name among the ones previously proposed, to each \gls{llm}.

We use zero-shot prompting for all the examples. Figure~\ref{fig:prompt_1} shows \textit{Prompt 1}, used to ask for name suggestions. We insert the corresponding rules with placeholders where \texttt{[rules]} is present.
The prompt is built in sections, separated by \texttt{\#\#\# section \#\#\#}:
in the beginning, we provide the context;
then, we give the instruction; 
lastly, we provide the desired output format.
The prompts used for the examples differ only in the predicates listed in the instruction and output format.
\review{Since we don't want our method to be dependent on the chosen language, we instruct the LLMs to behave as a ``specialist of logic programming''.}

\begin{figure}[h]
\centering
\begin{lstlisting}[breaklines=true]
### Context ###
You are a software engineer specialized in logic programming.

### Instruction ###
Given these first-order logic rules:
[rules]

Assign general and meaningful names to predicates h0, h1, h2, h3, and h4.
Do NOT change the body and the variables of the rules.
Give only ONE suggestion for each predicate.

### Output format ###
h0:
h1:
h2:
h3:
h4:
\end{lstlisting}
\caption{\textit{Prompt 1} for asking for one suggestion for each unnamed predicate. This prompt was used for Case Study~\ref{cs:family}. \texttt{[rules]} is replaced with the corresponding rules.}
\label{fig:prompt_1}
\end{figure}

\textit{Prompt 2}, shown in Figure~\ref{fig:prompt_2}, is used when asking the models to choose between their own suggestions. 
For each model, \texttt{[suggested names]} is replaced with the list of its own suggestions. 
Since models can suggest names with semantically equivalent meanings but written in different formats, such as using underscores, hyphens, or capitalized letters, we ensure that they are standardized: we remove any underscores or hyphens from the name and convert them to camel case, leaving the first letter in lowercase. 
After this, we remove duplicates from the list.

\begin{figure}[h] 
\centering
\begin{lstlisting}[breaklines=true]
### CONTEXT ###
You are a software engineer specialized in logic programming.
Given these first-order logic rules:
[rules]

The following names have been proposed for predicates h0, h1, h2, h3, and h4.
h0: [suggested names]
h1: [suggested names]
h2: [suggested names]
h3: [suggested names]
h4: [suggested names]

### INSTRUCTION ###
Choose the most suitable name for predicates h0, h1, h2, h3, and h4, among the proposed ones.

### OUTPUT FORMAT ###
h0: CHOSEN_NAME
h1: CHOSEN_NAME
h2: CHOSEN_NAME
h3: CHOSEN_NAME
h4: CHOSEN_NAME
\end{lstlisting}
\caption{\textit{Prompt 2} used for asking the models to choose one among their own suggestions. This prompt was used for Case Study~\ref{cs:family}. \texttt{[rules]} is replaced with the corresponding rules and the suggestions of the model being used replace \texttt{[suggested names]}.}
\label{fig:prompt_2}
\end{figure}

Finally, \textit{Prompt 3} in Figure~\ref{fig:prompt_3} was used for the judgment, where we asked the model to score the suggestions for each predicate. We also provide the scoring rules in the \texttt{INSTRUCTIONS} section.

\begin{figure}[h]
\centering
\begin{lstlisting}[breaklines=true]
### CONTEXT ###
You are a software engineer specialized in logic programming. 
The following logic rules have unnamed predicates in the head and/or in the body:
[rules]

These are the suggested names for the unnamed predicates h0, h1, h2, h3, and h4:
h0: [suggested names]
h1: [suggested names]
h2: [suggested names]
h3: [suggested names]
h4: [suggested names]

### INSTRUCTIONS ###
Score the proposed names for each predicate following the rules below:
- Assign 1 to correct and precise answers. 
- Assign 0.5 to answers that are too generic, but still correct.
- Assign 0 to incomplete, imprecise or incorrect answers.

### OUTPUT FORMAT ###
For each predicate, list all the suggestions and their score:
h0:
h1:
h2:
h3:
h4:
\end{lstlisting}
\caption{\textit{Prompt 3} used for asking the models acting as judges to score the suggestions for each predicate. This prompt was used for Case Study~\ref{cs:family}. \texttt{[rules]} is replaced with the corresponding rules.}
\label{fig:prompt_3}
\end{figure}

\subsection{LLMs used in this study}

For this study, we selected 7 different \glspl{llm}: 
OpenAI's ChatGPT based on GPT-4o (ChatGPT-4o) (\cite{openai2024gpt4ocard,openai2024hellogpt4o}) and ChatGPT based on o3mini (ChatGPT-o3mini) (\cite{openai2025gpto3minicard});
Meta's Llama 3.2 3B Instruct (Llama) (\cite{llama32intro,llama32mc}),
Google DeepMind's Gemini 1.5 Flash (Gemini) (\cite{gemini2024});
FalconMamba 7B Instruct (FalconMamba) (\cite{zuo2024falcon}) and Falcon3 7B Instruct (Falcon3) (\cite{falcon3}) of the \gls{tii};
and Cohere's Command R+ 08 2024 (Command R+) (\cite{cohere2024}).

ChatGPT is a powerful chatbot running on GPT-4o (with approximately 200 billion parameters), capable of processing multi-modal inputs and answering within a few milliseconds, mirroring the behavior of a human.
Generative pre-trained Transformer (GPT) models, based on the tranformers technology, are robust and versatile and are nowadays widely employed for routine tasks such as text summarization or translation. Recently, given the significant advances, the adoption of ChatGPT and \glspl{llm} in general is also spreading in more critical applications (\cite{koubaa2023exploring}), such as generating simplified clinical reports (\cite{jeblick2024chatgpt}) or supporting decision-making process (\cite{rao2023assessing}), with high accuracy and correctness.
In addition to GPT models, OpenAI provides also reasoning models, trained with reinforcement learning. 
Reasoning models, like o1 and the latest o3mini that has 3B parameters, use a chain-of-thought approach,
%to first think about the input and then deliver the answer. 
\review{where intermediate reasoning steps are generated before producing the final answer.}
Although this increases response time, reasoning models yield higher-quality outputs and are particularly effective in tasks involving science, mathematics, and coding (\cite{openai2025gpto3minicard}).

Google DeepMind's Gemini 1.5 Flash is a fast and versatile model that can take text, images, audio, and video as input, and produces a text output. Gemini 1.5 Flash has been adopted in an AI-based application to help users monitor and reduce their carbon emissions (\cite{dash2024enhancing}). It also obtained good results in answering multiple-choice questions regarding pediatric nephrology (\cite{mondillo2024basal}).

Llama 3.2 3B Instruct is a lightweight (3.21B parameters) text-only model of the Llama 3.2 family, which is a collection of multilingual, pre-trained, and instruction-tuned \glspl{llm}.

FalconMamba-7B-Instruction is a version of the base model FalconMamba 7B, fine-tuned on instruction data. The model is based on Mamba (\cite{gu2023mamba}), an attention-free architecture, and has 7.27B parameters. 
The Falcon3 family is a collection of pre-trained and instruction-tuned, decoder-only LLMs of various sizes.
Falcon3-7B-Instruct, specifically, has been pre-trained on curated high-quality and multilingual data from various sources (web, code, STEM), and according to \gls{tii} it achieves state-of-art-results on different tasks, including natural language processing, common-sense reasoning, coding, and reasoning. This is due to the improvements made in the development of the base models, such as depth upscaling and knowledge distillation (\cite{hinton2015distilling}).

Cohere's Command R+ 08-2024 is a multilingual model optimized for various applications, such as text summarization, reasoning, and structured data analysis, with 104 billion parameters, and trained on an extensive corpus of data in different languages. Command R+ 08-2024 has shown good performances for most of the writing tasks in Arabic tested by \cite{magdy2024gazelle} and evaluated on \textit{Gazelle}, the dataset they proposed, delivering clear and coherent responses.

\review{The experiments involving few-shot prompting, the \textit{reachability} case study, and the human judgment evaluation were conducted between July and August 2025, while all other experiments took place between January and February 2025. 
Since ChatGPT-4o and ChatGPT-o3mini were no longer available at the time of the \textit{reachability} case study experiment, we used ChatGPT-5, the only freely accessible version at that time.}

\subsection{Evaluation: LLM-as-a-Judge \review{and human judgment}}
We use \glspl{llm} also for the automated evaluation of the answers, adopting the LLM-as-a-Judge strategy (\cite{zheng2023judging,alammar2024hands}).

We selected four models, ChatGPT-4o, ChatGPT-o3mini, Gemini, and Command R+, to act as judges. Each judge assigns a score to every proposed name for each predicate: 1 for a correct and precise name, 0.5 for a name that is imprecise or too general but still correct, and 0 for a wrong name. The scores are then summed to determine the final results, and the name with the highest score is selected.

\review{
Finally, to assess the ability of \glspl{llm} in both renaming and judgment, we conducted a complementary experiment involving 14 human judges on all the case studies except Mutagenesis and \textit{reachability}. 
The \textit{reachability} case study was not included in the human judgment experiment, as it was introduced after the experiment had already been conducted, while Mutagenesis requires a field expert to provide reliable feedback.
Participants were asked to score the proposed names using the same evaluation method adopted by \glspl{llm} judges.
}

\section{Experiments and Results}\label{sec:exp}
In this section, we illustrate the results of our experiments.

For this study, the desktop app was used to execute both ChatGPT running on GPT-4o and ChatGPT running on GPT-o3mini.
We accessed Gemini and Command R+ via their own API.
For Llama, FalconMamba, and Falcon3, experiments were
performed locally via the Hugging Face Transformers framework (\cite{wolf2020transformers}) on a cluster consisting of two machines with two AMD EPYC 9654 96-cores and four machines with two AMD EPYC 9454 48-cores. The memory limit was set to 100 GB, and the time limit was set to 17 hours and 30 minutes.

\subsection{Prompting strategy}

%After multiple attempts, we settled on the input prompt shown in Figure \ref{fig:prompt_1}.
%In our prompt crafting approach, we initially provided relatively general instructions and then refined them based on the errors or undesired behaviors the models exhibited.
\review{We carried out several initial prompt design attempts to observe how LLMs responded under varying instructions.}
For instance, after noticing that the models often provided multiple equivalent suggestions, we decided to limit them to just one.
We also added a constraint to prevent models from altering the structure of the rules, after observing such behaviors with some \glspl{llm}.
\review{Each of these attempts was executed independently, ensuring that no contextual information was shared between runs. Importantly, the models did not retain any memory of previous executions; in particular, ChatGPT was used with the memory feature explicitly disabled to prevent any persistence of prior interactions. After analyzing the models’ responses and identifying recurring issues or undesired behaviors, we refined our instructions iteratively. The final version of the prompt, shown in Figure~\ref{fig:prompt_1}, was then used to rerun all experiments from scratch.}
%Additionally, due to the non-deterministic nature of the answers, and after observing that models sometimes gave the correct response right away but other times required several attempts, we decided to repeat the question a certain number of times to address this issue. In these preliminary experiments, we decided to use three repetitions.
Additionally, we observed that models sometimes gave the correct response on the first try, while other times they required several attempts. To address this, and given the non-deterministic nature of the answers, we decided to repeat the question a certain number of times to address this issue. In these preliminary experiments, we decided to use three repetitions.
%SPOSTATO SOPRA PER REVIEW We also added a constraint to prevent models from altering the structure of the rules, after observing such behaviors with some \glspl{llm}.
All the models % understood the task immediately
\review{handled the task appropriately,}
so zero-shot prompting was sufficient in this case. Most of them followed the instructions accurately, with only a few exceptions.
For example, Llama usually wrote a sentence (explaining the reasons for its choices) instead of adopting the requested format.
In such cases, however, the name can be extracted from the answer using basic string manipulation.
After getting three suggestions out of each model, we asked them to choose only one name among their own ones.

Tables \ref{tab:final_choices_A} and \ref{tab:final_choices_B} show the final choices made by each model for each predicate of each case study; we also report for each \gls{llm} how many of the suggested names could be considered correct.
\review{Table~\ref{tab:total_per_pred} provides a summary of the number of correct name suggestions per predicate. Each model receives a score of 1 if its final suggestion matches the expected name, and 0 otherwise. In the final column, we indicate how many \glspl{llm} correctly named each predicate. This overview allows us to easily compare the performance of the models on a per-predicate basis and identify which predicates were more challenging. 
}

\paragraph{Results for unnamed predicates appearing only in the head}
The \textit{coauthors} case study, which was the simplest, presented no difficulties for the models. In fact, all of them suggested a correct name, although sometimes too verbose.

\paragraph{Results for unnamed predicates appearing both in the head and in the body}
Regarding the \textit{family} case study, the GPT models and Command R+ provided the correct name of each predicate, and Gemini failed to identify only h4.
Llama was completely unable to find meaningful names for most of the predicates, with only two correct suggestions. Furthermore, one out of the three times it rewrote most of the rules, making its answer completely wrong.
FalconMamba and Falcon3 also struggled with the \textit{family} case study, as none of the answers were correct.
In particular, Falcon3 provided the answer for only one predicate, likely being misled by the word ``ONE'' in the prompt.
For this reason, its answers were not considered.

For the \textit{math} case study, the GPT models correctly identified almost all the names. 
Gemini followed closely, with only two mistakes.
Falcon3 and Command R+ also identified most of the correct names, while Llama just over half of them.
On the other hand, FalconMamba simply rewrote the rules with different capital letters all three times, and therefore its answers were ignored.
Many models couldn't find the correct name for predicates L (\texttt{isDivisor}, suggested only by Command R+), M (\texttt{gcd}), and N (\texttt{lcm}); these last two were correctly identified only by ChatGPT-4o and ChatGPT-o3mini.

\paragraph{Results for unnamed predicates appearing only in the body}
In the \textit{lcm} case study, most models struggled to identify the correct names for predicates G and H, namely \texttt{product} and \texttt{quotient} (or equivalent terms). 
Only ChatGPT-4o and ChatGPT-o3mini provided correct suggestions.

Finding \texttt{siblings} for predicate h3 in the \textit{cousins} case study was also challenging for some models, even if they suggested it in the \textit{family} case study. Only the GPT models, Gemini and LLama provided the correct name.

For the \textit{grandparent} case study, all the models correctly suggested \texttt{parent} for predicate h0, except for FalconMamba.

\begin{table}[h]
\scriptsize
\caption{Final names suggestions for each predicate of all the examples, made by ChatGPT-4o, ChatGPT-o3mini, and Command R+.}\label{tab:final_choices_A}%
\begin{tabular}{@{}lllll@{}}
\toprule
Predicate & ChatGPT-4o& ChatGPT-o3mini & Command R+\\

% coauthors
\midrule
\textit{\textbf{Coauthors}}\footnotemark[1] &&&\\
\textit{P} & coauthors\_with\_& coauthors & co\_authored\_paper\\
& common\_paper & & \\
\textbf{correct} & \textbf{1/1} & \textbf{1/1} & \textbf{1/1} \\

% family
\midrule
\textit{\textbf{Family}} &&&\\
\textit{h0} & parent & parent & parent \\
\textit{h1} & grandparent & grandparent & grandparent \\
\textit{h2} & common\_ancestor & common\_ancestor & common\_ancestor \\
\textit{h3} & sibling & sibling & sibling \\
\textit{h4} & cousin & cousin & cousins \\
\textbf{correct} & \textbf{5/5} & \textbf{5/5} & \textbf{5/5}\\

% grandparent 
\midrule
\textit{\textbf{Grandparent}} &&&\\
\textit{h0} & parent & parent & parent\\
\textbf{correct} & \textbf{1/1} & \textbf{1/1} & \textbf{1/1} \\

% cousins 
\midrule
\textit{\textbf{Cousins}} &&&\\
\textit{h3} & siblings & siblings & different\_parents \\
\textbf{correct} & \textbf{1/1} & \textbf{1/1} & \textbf{0/1} \\

% math 
\midrule
\textit{\textbf{Math}} &&&\\
\textit{A} & is\_number & numeric\_value/1 & is\_integer\\
\textit{B} & is\_even & even/1 & is\_even \\
\textit{C} & is\_odd & odd/1 & is\_odd \\
\textit{D} & absolute\_value & abs\_value/2 & negate\_if\_negative \\
\textit{E} & add & sum/3 & sum \\
\textit{F} & subtract & difference/3 & difference \\
\textit{G} & multiply & product/3 & product \\
\textit{H} & divide & quotient/3 & division \\
\textit{L} & is\_divisible & divides/2 & is\_divisor \\
\textit{M} & gcd & gcd/3 & modular\_arithmetic \\
\textit{N} & lcm & lcm/3 & nested\_calculation \\
\textit{P} & greater\_than & greater\_than/2 & greater\_than \\
\textit{Q} & greater\_than\_or\_equal & greater\_or\_equal/2 & greater\_equal \\
\textit{R} & less\_than & less\_than/2 & less\_than\\
\textit{S} & less\_than\_or\_equal & less\_or\_equal/2 & less\_equal \\
\textit{T} & equal & equal\_to/2 & equal \\
\textbf{correct} & \textbf{15/16} & \textbf{15/16} & \textbf{12/16} \\

% lcm 
\midrule
\textit{\textbf{lcm}} &&&\\
\textit{G} & compute\_product & multiply & is\_multiple\_of \\
\textit{H} & divide\_values & divide & lcm\_calculation \\
\textbf{correct} & \textbf{2/2} & \textbf{2/2} & \textbf{0/2} \\

% reachable
\midrule
\review{\textit{\textbf{reachability}}}&&&\\
\textit{inv1} & direct\_connection\footnotemark[2] & \textit{NA} & is\_connected\\
\textbf{correct} & \textbf{1/1} & \textbf{-} & \textbf{1/1}\\

\bottomrule
\end{tabular}
\footnotetext[1]{In this case, several names were suggested and even if excessively verbose, they can be considered correct.}
\review{\footnotetext[2]{Since ChatGPT-4o and ChatGPT-o3mini were no longer available at the time of the experiment, ChatGPT-5 was used instead.}}
\end{table}

\begin{table}[h]
\tiny
\caption{Final names suggestions for each predicate of all the examples, made by FalconMamba, Falcon3, Gemini, Llama.}\label{tab:final_choices_B}%
\begin{tabular}{@{}llllll@{}}
\toprule
Predicate & FalconMamba & Falcon3 & Gemini & Llama\\

% coauthors
\midrule
\textit{\textbf{Coauthors}}\footnotemark[1] &&&&\\
\textit{P} & Co-authoredResearchPaper & 
 CoAuthorResearchers & coauthored\_ & authoredTogether\\

&  &  & research\_paper & \\
\textbf{correct} & \textbf{1/1} & \textbf{1/1} & \textbf{1/1} & \textbf{1/1} \\

% family
\midrule
\textit{\textbf{Family}} &&&&\\
\textit{h0} & ancestor & - & parent & parent \\
\textit{h1} & ancestor & - & grandparent & ancestor \\
\textit{h2} & ancestor & - & commonAncestor & great\_ancestor \\
\textit{h3} & sister & - & sibling & sibling\\
\textit{h4} & h3 & - & halfSibling & full\_sibling \\
\textbf{correct} & \textbf{0/5} & \textbf{0/5} & \textbf{4/5} & \textbf{2/5} \\

% grandparent 
\midrule
\textit{\textbf{Grandparent}} &&&&\\
\textit{h0} & rename\_h0 & parent & parent & parent\\
\textbf{correct} & \textbf{0/1} & \textbf{1/1} & \textbf{1/1} & \textbf{1/1} \\

% cousins 
\midrule
\textit{\textbf{Cousins}} &&&&\\
\textit{h3} & is\_third\_degree\_relative & parent & sibling & sibling\\
\textbf{correct} & \textbf{0/1} & \textbf{0/1} & \textbf{1/1} & \textbf{1/1} \\

% math 
\midrule
\textit{\textbf{Math}} &&&&\\
\textit{A} & - & IsNumber & isNumber & is\_integer \\
\textit{B} & - & IsEven & isEven & is\_odd \\
\textit{C} & - & IsOdd & isOdd & is\_even \\
\textit{D} & - & IsPositiveDifference & absValue & next \\
\textit{E} & - & Sum & sum & add \\
\textit{F} & - & Difference & difference & subtract \\
\textit{G} & - & Product & product & multiply \\
\textit{H} & - & Division & quotient & divide \\
\textit{L} & - & DivisibleBy & isDivisibleBy & modulo \\
\textit{M} & - & CustomOperation & gcd & accumulate \\
\textit{N} & - & ComplexOperation & complexCalculation & accumulate\_sequence \\
\textit{P} & - & GreaterThan & greaterThan & greater\_than \\
\textit{Q} & - & GreaterThanOrEqual & greaterThanOrEqual & greater\_than\\
 &  &  &  & or\_equal\_to \\
\textit{R} & - & LessThan & lessThan & less\_than \\
\textit{S} & - & LessThanOrEqual & lessThanOrEqual & less\_than\_\\
 &  &  &  & or\_equal\_to \\
\textit{T} & - & Equal & isEqual & equal\_to \\
\textbf{correct} & \textbf{0/16} & \textbf{12/16} & \textbf{14/16} & \textbf{9/16} \\

% lcm 
\midrule
\textit{\textbf{lcm}} &&&&\\
\textit{G} & lowest\_common\_multiple & multiply\_and\_add & find\_least\_common\_ & greatest\_\\

& & & multiple\_intermediate & common\_divisor\\

\textit{H} & highest\_common\_divisor & divide\_and\_subtract & compute\_lcm\_ & least\_\\

& &  & from\_gcd & common\_multiple\\
\textbf{correct} & \textbf{0/2} & \textbf{0/2} & \textbf{0/2} & \textbf{0/2}\\

% reachable
\midrule
\review{\textit{\textbf{reachability}}} &&&\\
\textit{inv1} & can reach & - & directly\_connected & is\_connected\\
\textbf{correct} & \textbf{0/1} & \textbf{0/1} & \textbf{1/1} & \textbf{1/1}\\

\bottomrule
\end{tabular}
\footnotetext[1]{In this case, several names were suggested and even if excessively verbose, they can be considered correct.}
\end{table}

\begin{table}[h]
%\tiny
\caption{\review{Summary of correct name suggestions for each predicate. For each LLM, a value of 1 indicates that its final suggestion was correct, while 0 indicates an incorrect suggestion. The \textit{Total} column shows the number of LLMs that correctly named each predicate.}}\label{tab:total_per_pred}
% \begin{tabular}{@{}lcccccccc@{}}
\begin{tabularx}{\textwidth}{ 
  >{\raggedright\arraybackslash}X
  >{\centering\arraybackslash}X
  >{\centering\arraybackslash}X
  >{\centering\arraybackslash}X
  >{\centering\arraybackslash}X
  >{\centering\arraybackslash}X
  >{\centering\arraybackslash}X
  >{\centering\arraybackslash}X
  >{\centering\arraybackslash}X
  }
\toprule
Predicate & ChatGPT & ChatGPT & Command & Falcon & Falcon3 & Gemini & Llama & \textbf{Total}\\
 & 4o & o3mini & R+ & Mamba &  &  &  & \\

% coauthors
\midrule
\textit{\textbf{Coauthors}} &&&&&\\
\textit{P} & 1 & 1 & 1 & 1 & 1 & 1 & 1 & \textbf{7/7}\\

% family
\midrule
\textit{\textbf{Family}} &&&&&\\
\textit{h0} & 1 & 1 & 1 & 0 & 0 & 1 & 1 & \textbf{5/7}\\
\textit{h1} & 1 & 1 & 1 & 0 & 0 & 1 & 0 & \textbf{4/7}\\
\textit{h2} & 1 & 1 & 1 & 0 & 0 & 1 & 0 & \textbf{4/7}\\
\textit{h3} & 1 & 1 & 1 & 0 & 0 & 1 & 1 & \textbf{5/7}\\
\textit{h4} & 1 & 1 & 1 & 0 & 0 & 0 & 0 & \textbf{3/7}\\

% grandparent 
\midrule
\textit{\textbf{Grandparent}} &&&&&\\
\textit{h0} & 1 & 1 & 1 & 0 & 1 & 1 & 1 & \textbf{6/7}\\

% cousins 
\midrule
\textit{\textbf{Cousins}} &&&&&\\
\textit{h3} & 1 & 1 & 0 & 0 & 0 & 1 & 1 & \textbf{4/7}\\

% math 
\midrule
\textit{\textbf{Math}} &&&&\\
\textit{A} & 1 & 1 & 0 & 0 & 1 & 1 & 0 & \textbf{4/7}\\
\textit{B} & 1 & 1 & 1 & 0 & 1 & 1 & 0 & \textbf{5/7}\\
\textit{C} & 1 & 1 & 1 & 0 & 1 & 1 & 0 & \textbf{5/7}\\
\textit{D} & 1 & 1 & 0 & 0 & 0 & 1 & 0 & \textbf{3/7}\\
\textit{E} & 1 & 1 & 1 & 0 & 1 & 1 & 1 & \textbf{6/7}\\
\textit{F} & 1 & 1 & 1 & 0 & 1 & 1 & 1 & \textbf{6/7}\\
\textit{G} & 1 & 1 & 1 & 0 & 1 & 1 & 1 & \textbf{6/7}\\
\textit{H} & 1 & 1 & 1 & 0 & 1 & 1 & 1 & \textbf{6/7}\\
\textit{L} & 0 & 0 & 1 & 0 & 0 & 0 & 0 & \textbf{1/7}\\
\textit{M} & 1 & 1 & 0 & 0 & 0 & 1 & 0 & \textbf{3/7}\\
\textit{N} & 1 & 1 & 0 & 0 & 0 & 0 & 0 & \textbf{2/7}\\
\textit{P} & 1 & 1 & 1 & 0 & 1 & 1 & 1 & \textbf{6/7}\\
\textit{Q} & 1 & 1 & 1 & 0 & 1 & 1 & 1 & \textbf{6/7}\\
\textit{R} & 1 & 1 & 1 & 0 & 1 & 1 & 1 & \textbf{6/7}\\
\textit{S} & 1 & 1 & 1 & 0 & 1 & 1 & 1 & \textbf{6/7}\\
\textit{T} & 1 & 1 & 1 & 0 & 1 & 1 & 1 & \textbf{6/7}\\

% lcm 
\midrule
\textit{\textbf{lcm}} &&&&\\
\textit{G} & 1 & 1 & 0 & 0 & 0 & 0 & 0 & \textbf{2/7}\\
\textit{H} & 1 & 1 & 0 & 0 & 0 & 0 & 0 & \textbf{2/7}\\

% reachable
\midrule
\review{\textit{\textbf{reachability}}} &&&&&\\
\textit{inv1} & 1\footnotemark[1] & - & 1 & 0 & 0 & 1 & 1 & \textbf{4/6}\\

\bottomrule
\end{tabularx}
\footnotetext[1]{Since ChatGPT-4o and ChatGPT-o3mini were no longer available at the time of the experiment, ChatGPT-5 was used instead.}
\end{table}

\paragraph{Results for the Mutagenesis case study}

In Table \ref{tab:muta_results} we report the results obtained on the Mutagenesis ring theory, using ChatGPT-4o and ChatGPT-o3mini.
Some of the predicates appear to be similar to the original ones, but overall it is difficult to determine whether the suggested names are meaningful. In such a case, a domain expert should make the final decision. 
To improve the quality of the output, the models could be fine-tuned on the specific domain; alternatively, context (e.g., in the form of scientific papers) could be provided.

It is worth mentioning another attempt we made with Mutagenesis. For this initial experiment, we used a less constrained version of \textit{Prompt 1}, where we simply asked for suggestions for all the predicates, without specifying the context or limiting the model to provide only one suggestion per predicate.
The response of Command R+ was particularly interesting. The model not only explained the predicate to rename but in some cases it also invented a helper predicate for it.
An example of such behavior is shown in Figure \ref{fig:muta_pi}: the model created the helper predicate  \texttt{check\_atom\_count\_recursive} for predicate HP19, which was renamed as \texttt{check\_atom\_count\_recursive}. Regardless of the correctness of the suggested name and despite altering the structure of the rule, this suggests the potential of \glspl{llm} to directly perform \gls{pi}. 

\begin{figure}[h]
\centering
\begin{lstlisting}[breaklines=true]
% Rules for specific chemical patterns

% HP19: Checks if a drug contains a specific atom type with a certain number of occurrences.
% Args: Drug, Number of atoms, List of atoms, List of atom types
check_atom_count(Drug, N, Atoms, Types) :- 
   atm(Drug, Atom, Type, _, _),
   Types = [Type | _],
   check_atom_count_recursive(Drug, N, [Atom], Atoms, Types, 1).

check_atom_count_recursive(_, 1, Atoms, Atoms, _, _) :- !. % Base case: N is 1
check_atom_count_recursive(Drug, N, [Atom | _], Atoms, [Type | Types], N) :-
   N > 1,
   N1 is N - 1,
   atm(Drug, Atom, Type, _, _),
   check_atom_count_recursive(Drug, N1, _, Atoms, Types, _).
\end{lstlisting}
\caption{Result obtained with Command R+ on Case Study~\ref{cs:muta_example_rules} with a less constrained version of \textit{Prompt 1}.}
\label{fig:muta_pi}
\end{figure}

\begin{table}[h]
\caption{Results obtained on the Mutagenesis ring theory, with ChatGPT-4o and ChatGPT-o3mini.}
\label{tab:muta_results}
\centering
\scriptsize
\begin{tabular}{@{}lll@{}}
\toprule
Predicate & ChatGPT-4o & ChatGPT-o3mini\\
\midrule

HP1 (anthracene) & fused\_aromatic\_ & fused\_aromatic\_tricyclic\_\\
&tricycles\_non\_overlapping&system\_no\_bridge\\

HP2 (phenanthrene) & fused\_aromatic\_ & fused\_aromatic\_tricyclic\_\\
&tricycles\_adjacent&system\_with\_bridge\\

HP3 (ball3) & fused\_aromatic\_ & fused\_aromatic\_tricyclic\_\\
&tricycles\_interconnected&system\_common\_junction\\

HP4 (members\_bonded) & shared\_aromatic\_bond & ring\_junction\_bridge\\

HP5 (no\_of\_benzenes) & aromatic\_hexagons\_count & count\_aromatic\_\\
&&six\_membered\_rings\\

HP6 (no\_of\_carbon\_5\_ & aromatic\_pentagons\_count & count\_aromatic\_\\
aromatic\_rings)&&five\_membered\_rings\\

HP7 (no\_of\_carbon\_6\_rings) & non\_benzenoid\_ & count\_nonaromatic\_\\
&hexagons\_count&six\_membered\_rings\\

HP8 (no\_of\_carbon\_5\_rings) & non\_benzenoid\_ & count\_nonaromatic\_\\
&pentagons\_count&five\_membered\_rings\\

HP9 (no\_of\_hetero\_ & heterocyclic\_hexagons & count\_aromatic\_heterocyclic\\
aromatic\_6\_rings)&\_count&\_six\_membered\_rings\\

HP10 (no\_of\_hetero\_ & heterocyclic\_pentagons\_count & count\_aromatic\_heterocyclic\\
aromatic\_5\_rings)&\_count &five\_membered\_rings\\

HP11 (ring\_size\_6) & hexagonal\_ring\_structure & generic\_six\_\\
&&membered\_ring\\

HP12 (ring\_size\_5) & pentagonal\_ring\_structure & generic\_five\_membered\_ring\\

HP13 (benzene) & benzene\_ring & aromatic\_six\_membered\_ring\\

HP14 (carbon\_5\_ & cyclopentadiene\_ring & aromatic\_five\_\\
aromatic\_ring)&&membered\_ring\\

HP15 (carbon\_6\_ring) & non\_benzenoid\_hexagon & nonaromatic\_six\_\\
&&membered\_ring\\

HP16 (carbon\_5\_ring) & non\_benzenoid\_pentagon & nonaromatic\_\\
&&five\_membered\_ring\\

HP17 (hetero\_ & heterocyclic\_hexagon & aromatic\_heterocyclic\_\\
aromatic\_6\_ring)&&six\_membered\_ring\\

HP18 (hetero\_ & heterocyclic\_pentagon & aromatic\_heterocyclic\_\\
aromatic\_5\_ring)&&five\_membered\_ring\\

HP19 (atoms) & atom\_sequence & ring\_atom\_chain\\

HP20 (ring6) & aromatic\_hexagonal\_ & six\_membered\_\\
&bond\_pattern&ring\_structure\\

HP21 (ring5) & aromatic\_pentagonal\_ & five\_membered\_\\
&bond\_pattern&ring\_structure\\

HP22 (no\_of\_nitros) & nitro\_group\_count & count\_nitro\_groups\\
HP23 (nitro) & nitro\_group\_structure & nitro\_group\_structure\\
HP24 (methyl) & methyl\_group\_structure & methyl\_group\_structure\\

\bottomrule
\end{tabular}
\end{table}

\review{
\paragraph{Results for the reachability case study}
The \textit{reachability} case study was generally unchallenging for the majority of models evaluated. 
%GPT-5, Command R+, Gemini, and Llama each produced plausible and appropriate suggestions. 
Suggestions produced by GPT-5 (direct\_connection), Command R+ (is\_connected), Gemini (directly\_connected), and LLama (is\_connected) are all plausible and appropriate, as they generalize a single-step path relation.
In contrast, both FalconMamba and Falcon3 exhibited notable difficulties. FalconMamba generated an output (can reach) that was syntactically invalid as a predicate, due to the inclusion of a space in the identifier, while Falcon 3 failed to produce any answer at all.
}

\review{
\subsection{Results for the few-shot prompting experiment}
We additionally conducted a few-shot prompting experiment with the \glspl{llm} that demonstrated lower performance, namely FalconMamba, Falcon3, and Llama, with the goal of improving their answers. 
We focused on the \textit{math} case study, which was selected due to its potential to highlight the benefits of few-shot prompting. 
In this experiment, suggestions were requested for one predicate at a time, while also providing the model with the correct names of previously addressed predicates. 
Figure~\ref{fig:fewshot} shows the prompts used in the few-shot prompting experiments with answers from Falcon3. The initial prompt, \textit{Prompt 0}, requests the name of only the first predicate (\textit{A}). Then, its answers is used to replace the corresponding unnamed predicate in the following prompt, \textit{Prompt 1}, while requesting a suggestion for the second predicate (\textit{P}). This continues until the last prompt, in which all the unnamed predicates of the math case study have been renamed, except of the last one.
This incremental approach aimed to support the model by narrowing its focus and reducing the complexity of the task, in contrast to requiring it to generate names for all predicates simultaneously.
However, the outcomes did not meet expectations. None of the models showed improved performance in terms of correctly identifying a greater number of names; in some instances, performance even declined.
}

\begin{figure}
\scriptsize
\textbf{Prompt 0:}
\begin{lstlisting}[breaklines=true]
### CONTEXT ###
You are a software engineer specialized in logic programming.

### INSTRUCTION ###
Given the logic rules below, find a meaningful name for A.
Give only ONE suggestion for the predicate name.
Write only the predicate name after the predicate placeholder, without any additional text. 
Do not rewrite the predicate body.

A(X) :- integer(X), !.
A(X) :- float(X).

Strictly follow this output format:
A: [your_suggestion]
\end{lstlisting}

\textbf{Response:}
\begin{lstlisting}[breaklines=true]
A: is_number
\end{lstlisting}

\textbf{Prompt 1:}
\begin{lstlisting}[breaklines=true]
### CONTEXT ###
You are a software engineer specialized in logic programming.

### INSTRUCTION ###
Given the logic rules below, find a meaningful name for P.
Give only ONE suggestion for the predicate name.
Write only the predicate name after the predicate placeholder, without any additional text. 
Do not rewrite the predicate body.

is_number(X) :- integer(X), !.
is_number(X) :- float(X).

P(X,Y) :- is_number(X), is_number(Y), X > Y.

Strictly follow this output format:
P: [your_suggestion]
\end{lstlisting}

\textbf{Response:}
\begin{lstlisting}[breaklines=true]
P: is_greater_number
\end{lstlisting}

\caption{\review{Prompts used in the few-shot prompting experiments with Falcon3 answers.}}
\label{fig:fewshot}
\end{figure}

\review{
Table~\ref{tab:fewshot} shows the results of this experiment.
With few-shot prompting Falcon3 correctly identified 10 out of 16 predicate names, a decrease from the 12 out of 16 correctly identified with zero-shot prompting. 
Similarly, Llama achieved 8 correct predictions with few-shot prompting, compared to 9 without. 
%It is worth noting that FalconMamba generated some suggestions following few-shot prompting, four compared to zero with zero-shot prompting, although only two of these were deemed acceptable.
%These results suggest that few-shot prompting, as applied in this setting, did not lead to improved performance for these models, however, breaking down the prompt and using few-shot prompting may benefit certain models.
A substantial improvement was observed with FalconMamba. with zero-shot prompting, the model failed to generate a meaningful output, simply repeating the predicates.
In contrast, when few-shot prompting was applied, the model provided a suggestion for all the predicates, finding 11 correct answers out of 16. 
These findings, particularly in the case of FalconMamba, indicate that breaking down the prompt and using few-shot prompting can significantly enhance performance in models that otherwise struggle with zero-shot prompting.
}

\begin{table}[h]
\caption{\review{FalconMamba, Falcon3 and Llama suggestions for the predicates of the \textit{math} case study, in the few-shot prompting experiment.}}
\label{tab:fewshot}
\centering
\footnotesize
\begin{tabular}{@{}llll@{}}
\toprule
Predicate & FalconMamba & Falcon3 & Llama\\
\midrule
A : number & integer\_or\_float & is\_number & has\_integer\_value\\
B : even & even\_integer & is\_even\_number & parity\\
C : odd & odd\_integer & is\_odd\_number & odd\\
D : abs & abs & negate\_if\_negative & divide\_by\_zero\\
E : addition & sum & add\_numbers & sum\\
F : substraction & subtract & subtract\_numbers & difference\\
G : multiplication & multiplication & multiply\_numbers & multiply\\
H : division & division\_by\_non\_zero\_integer & divide\_numbers & divide\\
L : divisor & even\_integer & divisible\_by & is\_divisible\\
M : gdc & even\_integer & swap\_and\_mod\_numbers & min\\
N : lcm & even\_integer\_product & calculate\_result & number\\
P : greater\_than & greater\_than\_or\_equal & is\_greater\_number & greater\_integer\\
Q : greater\_equal\_than & greater\_than\_or\_equal & is\_number\_or\_equal & greater\_than\\
R : less\_than & less\_than & is\_less\_number & less\_than\\
S : less\_equal\_than & less\_than & is\_number\_or\_less & between\\
T : equal & equal\_to & is\_equal\_number & equal\\
\textbf{correct} & \textbf{11/16} & \textbf{10/16} & \textbf{8/16}\\
\bottomrule
\end{tabular}
\end{table}

\subsection{LLMs judgment results}
Regarding the judgment on the \textit{coauthors} case study, shown in Table \ref{tab:coauth_gp_cous_lcm_results}, all the suggestions 
%can be considered correct
are plausible. \gls{llm} judges seem to prefer a longer and more detailed name, assigning the lowest score to the simplest one, i.e. \texttt{coauthors}.

In Table \ref{tab:family_results} we report the results of the judgment on the \textit{family} case study.
The correct name was chosen for each predicate.
The same holds for the \textit{grandparent} and \textit{cousins} case studies, both shown in Table \ref{tab:coauth_gp_cous_lcm_results}.

Table \ref{tab:math_results} shows the results of the judgment on the \textit{math} case study. Here, the correct name was chosen for all the predicates, except for predicate L. While the correct name is \texttt{divisor}, it was mostly interpreted as ``divisible'', which is actually the opposite.
Regarding predicates M and N, all the judges recognized the true names, \texttt{gcd} and \texttt{lcm} respectively, even if they didn't suggest them initially. The strategy of asking multiple models for name suggestions turned out to be effective, as it allowed the judges to identify the correct name when one model suggested it, even when the others did not.
The same cannot be said for the \textit{lcm} case study, as can be seen in Table \ref{tab:coauth_gp_cous_lcm_results}. For predicate G, \texttt{findLeastCommonMultipleIntermediate} was chosen as the most fitting name, instead of \texttt{computeProduct} or \texttt{multiply}, which would have been equally correct.
Similarly, for predicate H, \texttt{computeLcmFromGcd} was selected as the most appropriate name, rather than \texttt{divide} or \texttt{divideValues}.

\review{
Table~\ref{tab:reachable_jdg_results} presents the judgment results for the reachability case study. It is important to note that ChatGPT-5 was used as a substitute for ChatGPT-4o and ChatGPT-o3mini, as these models were no longer available at the time the experiment was conducted.
Three out of the four alternatives were selected as the most suitable name, each receiving the same score. This outcome is acceptable, as all three options are considered plausible. Notably, however, only ChatGPT-5 assigned a score of 0 to the proposed name \texttt{can reach}, which is syntactically incorrect. In contrast, both Gemini and Command R+ assigned it a score of 0.5, indicating a limited ability to recognize the syntactic invalidity of the suggestion.
}

\begin{table}[h]
\caption{\review{LLM judgment results for Case Studies \ref{cs:coauthors} (\textit{coauthors}) and \ref{cs:gp_cs_lcm} (\textit{grandparent}, \textit{cousins} and \textit{lcm}). Judges assign a score of 1 for correct and precise names, 0.5 for imprecise or overly general names that are still correct, and 0 for incorrect names. The final score, shown in the last column, is the \review{average} of the individual scores given by the judges. Names that can be considered correct are marked with an asterisk (``*'').
}}
\label{tab:coauth_gp_cous_lcm_results}
\centering
\scriptsize
\begin{tabular}{@{}lccccc@{}}
\toprule
\textbf{Case Study} - \textit{Predicate} & ChatGPT & ChatGPT & Gemini & Command & \textbf{\textbf{Score}} \\
& (4o) & (o3mini) && R+ & \\
\midrule
\textbf{coauthors} - \textit{P}: & & & & &\\
coauthoredResearchPaper* & 1 & 1 & 1 & 1 & \textbf{1.000}\\
coauthorsWith- & 1 & 1 & 1 & 1 & \textbf{1.000}\\
CommonPaper* &&&&&\\
coAuthoredPaper* & 0.5 & 1 & 0.5 & 1 & \textbf{0.750}\\
coAuthorResearchers* & 1 & 0.5 & 0.5 & 1 & \textbf{0.750}\\
authoredTogether* & 0.5 & 0.5 & 0.5 & 1 & \textbf{0.625}\\
coauthors* & 0.5 & 0.5 & 0.5 & 0.5 & \textbf{0.500}\\
\hline
\textbf{grandparent} - \textit{h0}: & & & & &\\
parent* & 1 & 1 & 1 & 1 & \textbf{1.000}\\
renameH0 & 0 & 0 & 0 & 0.5 & \textbf{0.125}\\
\hline
\textbf{cousins} - \textit{h3}: & & & & &\\
siblings* & 1 & 1 & 1 & 0 & \textbf{0.750}\\
sibling* & 1 & 1 & 0 & 0 & \textbf{0.500}\\
differentParents & 0 & 0 & 1 & 1 & \textbf{0.500}\\
isThirdDegreeRelative & 0.5 & 0 & 0.5 & 0 & \textbf{0.250}\\
parent & 0 & 0 & 0 & 0 & \textbf{0.000}\\

\midrule
\textbf{lcm} - \textit{G} & & & & & \\
findLeastCommon- & 1 & 0.5 & 1 & 0 & \textbf{0.625} \\
MultipleIntermediate &&&&&\\
computeProduct* & 0.5 & 1 & 0.5 & 0 & \textbf{0.500} \\
multiply* & 0.5 & 1 & 0.5 & 0 & \textbf{0.500} \\
greatestCommonDivisor & 0 & 0 & 0 & 1 & \textbf{0.250} \\
isMultipleOf & 0 & 0 & 0 & 0 & \textbf{0.000} \\
lowestCommonMultiple & 0 & 0 & 0 & 0 & \textbf{0.000} \\
multiplyAndAdd & 0 & 0 & 0 & 0 & \textbf{0.000}\vspace{0.5em}\\

\textbf{lcm} - \textit{H} & & & & &\\
computeLcmFromGcd & 1 & 1 & 1 & 0.5 & \textbf{0.875} \\
divide* & 0.5 & 1 & 0.5 & 0.5 & \textbf{0.625} \\
lcmCalculation & 1 & 0.5 & 0.5 & 0.5 & \textbf{0.625} \\
divideValues* & 0.5 & 1 & 0.5 & 0 & \textbf{0.500} \\
leastCommonMultiple & 0 & 0 & 0 & 1 & \textbf{0.250} \\
highestCommonDivisor & 0 & 0 & 0 & 0 & \textbf{0.000} \\
divideAndSubtract & 0 & 0 & 0 & 0 & \textbf{0.000} \\
\bottomrule
\end{tabular}
\end{table}

\begin{table}[h]
\caption{LLM judgment results for Case Study \ref{cs:family} (\textit{family}). Judges assign a score of 1 for correct and precise names, 0.5 for imprecise or overly general names that are still correct, and 0 for incorrect names. The final score, shown in the last column, is the \review{average} of the individual scores given by the judges. Names that can be considered correct are marked with an asterisk (``*'').}
\label{tab:family_results}
\centering
\scriptsize
\begin{tabular}{@{}lccccc@{}}
\toprule
Predicate & ChatGPT & ChatGPT & Gemini & Command & \textbf{\textbf{Score}} \\
& (4o) & (o3mini) && R+ & \\
\midrule
\textit{h0} & & & & & \\
parent* & 1 & 1 & 1 & 1 & \textbf{1.000} \\
ancestor & 0.5 & 0 & 0.5 & 0.5 & \textbf{0.375} \\
\hline
\textit{h1} & & & & & \\
grandparent* & 1 & 1 & 1 & 1 & \textbf{1.000} \\
ancestor & 0.5 & 0.5 & 0.5 & 0.5 & \textbf{0.500} \\
\hline
\textit{h2} & & & & & \\
commonAncestor* & 1 & 1 & 1 & 0.5 & \textbf{0.875} \\
ancestor & 0.5 & 0.5 & 0.5 & 0 & \textbf{0.375} \\
greatAncestor & 0.5 & 0 & 0 & 0.5 & \textbf{0.250} \\
\hline
\textit{h3} & & & & & \\
sibling* & 1 & 1 & 1 & 1 & \textbf{1.000} \\
sister & 0.5 & 0 & 0.5 & 0.5 & \textbf{0.375} \\
\hline
\textit{h4} & & & & & \\
cousin* & 1 & 1 & 1 & 0.5 & \textbf{0.875} \\
cousins* & 1 & 1 & 0.5 & 0.5 & \textbf{0.750} \\
halfSibling & 0 & 0 & 0.5 & 0.5 & \textbf{0.250} \\
fullSibling & 0 & 0 & 0.5 & 0.5 & \textbf{0.250} \\
h3 & 0 & 0 & 0 & 0 & \textbf{0.000}\\
\bottomrule
\end{tabular}
\end{table}

\begin{table}[h]
\caption{LLM judgment results for Case Study \ref{cs:math} (\textit{math}). Judges assign a score of 1 for correct and precise names, 0.5 for imprecise or overly general names that are still correct, and 0 for incorrect names. The final score, shown in the last column, is the \review{average} of the individual scores given by the judges. Names that can be considered correct are marked with an asterisk (``*'').}
\label{tab:math_results}
\centering
\scriptsize
\begin{tabular}{@{}lccccc@{}}
\toprule
Predicate & ChatGPT & ChatGPT & Gemini & Command & \textbf{\textbf{Score}} \\
& (4o) & (o3mini) && R+ & \\
\midrule
\textit{A} & & & & & \\
isNumber* & 1 & 1 & 1 & 1 & \textbf{1.000} \\
numericValue* & 0.5 & 1 & 0.5 & 0.5 & \textbf{0.625} \\
isInteger & 0 & 0 & 0.5 & 0.5 & \textbf{0.250} \\
\hline
\textit{B} & & & & & \\
isEven* & 1 & 1 & 1 & 1 & \textbf{1.000} \\
even* & 1 & 1 & 1 & 0.5 & \textbf{0.875} \\
isOdd & 0 & 0 & 0 & 0 & \textbf{0.000} \\
\hline
\textit{C} & & & & & \\
isOdd* & 1 & 1 & 1 & 1 & \textbf{1.000} \\
odd* & 1 & 1 & 1 & 0.5 & \textbf{0.875} \\
isEven & 0 & 0 & 0 & 0 & \textbf{0.000} \\
\hline
\textit{D} & & & & & \\
absoluteValue* & 1 & 1 & 1 & 1 & \textbf{1.000} \\
absValue* & 1 & 1 & 1 & 0.5 & \textbf{0.875} \\
negateIfNegative & 1 & 0.5 & 1 & 0.5 & \textbf{0.750} \\
isPositiveDifference & 0 & 0 & 0.5 & 0 & \textbf{0.125} \\
next & 0 & 0 & 0 & 0 & \textbf{0.000} \\
\hline
\textit{E} & & & & & \\
add* & 1 & 1 & 1 & 1 & \textbf{1.000} \\
sum* & 1 & 1 & 1 & 0.5 & \textbf{0.875} \\
\hline
\textit{F} & & & & & \\
subtract* & 1 & 1 & 1 & 1 & \textbf{1.000} \\
difference* & 1 & 0.5 & 1 & 0.5 & \textbf{0.750} \\
\hline
\textit{G} & & & & & \\
multiply* & 1 & 1 & 1 & 1 & \textbf{1.000} \\
product* & 1 & 1 & 1 & 0.5 & \textbf{0.875} \\
\hline
\textit{H} & & & & & \\
divide* & 1 & 1 & 1 & 1 & \textbf{1.000} \\
division* & 1 & 1 & 0.5 & 0.5 & \textbf{0.750} \\
quotient* & 1 & 0.5 & 1 & 0.5 & \textbf{0.750} \\
\hline
\textit{L} & & & & & \\
divides & 1 & 1 & 0.5 & 0.5 & \textbf{0.750} \\
divisibleBy & 1 & 0.5 & 1 & 0.5 & \textbf{0.750} \\
isDivisible & 1 & 0 & 1 & 1 & \textbf{0.750} \\
isDivisibleBy & 1 & 0.5 & 1 & 0.5 & \textbf{0.750} \\
isDivisor* & 0.5 & 1 & 0 & 0.5 & \textbf{0.500} \\
modulo & 0.5 & 0 & 0.5 & 0.5 & \textbf{0.375} \\
\hline
\textit{M} & & & & & \\
gcd* & 1 & 1 & 1 & 1 & \textbf{1.000} \\
accumulate & 0 & 0 & 0.5 & 0.5 & \textbf{0.250} \\
modularArithmetic & 0 & 0 & 0.5 & 0.5 & \textbf{0.250} \\
customOperation & 0 & 0 & 0 & 0.5 & \textbf{0.125} \\
\hline
\textit{N} & & & & & \\
lcm* & 1 & 1 & 1 & 1 & \textbf{1.000} \\
complexCalculation & 0 & 0.5 & 0.5 & 0.5 & \textbf{0.375} \\
complexOperation & 0 & 0.5 & 0.5 & 0.5 & \textbf{0.375} \\
nestedCalculation & 0 & 0.5 & 0.5 & 0.5 & \textbf{0.375} \\
accumulateSequence & 0 & 0 & 0.5 & 0.5 & \textbf{0.250} \\
\hline
\textit{P} & & & & & \\
greaterThan* & 1 & 1 & 1 & 1 & \textbf{1.000} \\
\hline
\textit{Q} & & & & & \\
greaterThanOrEqual* & 1 & 1 & 1 & 1 & \textbf{1.000} \\
greaterEqual* & 1 & 1 & 1 & 0.5 & \textbf{0.875} \\
greaterOrEqual* & 1 & 1 & 1 & 0.5 & \textbf{0.875} \\
greaterThanOrEqualTo* & 1 & 1 & 1 & 0.5 & \textbf{0.875} \\
\hline
\textit{R} & & & & & \\
lessThan* & 1 & 1 & 1 & 1 & \textbf{1.000} \\
\hline
\textit{S} & & & & & \\
lessThanOrEqual* & 1 & 1 & 1 & 1 & \textbf{1.000} \\
lessEqual* & 1 & 1 & 1 & 0.5 & \textbf{0.875} \\
lessOrEqual* & 1 & 1 & 1 & 0.5 & \textbf{0.875} \\
lessThanOrEqualTo* & 1 & 1 & 1 & 0.5 & \textbf{0.875} \\
\hline
\textit{T} & & & & & \\
equal* & 1 & 1 & 1 & 1 & \textbf{1.000} \\
equalTo* & 1 & 1 & 1 & 0.5 & \textbf{0.875} \\
isEqual* & 1 & 1 & 1 & 0.5 & \textbf{0.875} \\
\bottomrule
\end{tabular}
\end{table}

% \begin{table}[h]
% \caption{LLM judgment results for the \textit{lcm} case study. Judges assign a score of 1 for correct and precise names, 0.5 for imprecise or overly general names that are still correct, and 0 for incorrect names. The final score, shown in the last column, is the \review{average} of the individual scores given by the judges. Names that can be considered correct are marked with an asterisk (``*'').}
% \label{tab:lcm_results}
% \centering
% \scriptsize
% \begin{tabular}{@{}lccccc@{}}
% \toprule
% \textbf{Predicate} & \textbf{ChatGPT} & \textbf{ChatGPT} & \textbf{Gemini} & \textbf{Command} & \textbf{\textbf{Score}} \\
% & \textbf{(4o)} & \textbf{(o3mini)} && R+ & \\
% \midrule
% \textit{G} & & & & & \\
% findLeastCommon- & 1 & 0.5 & 1 & 0 & \textbf{2.5} \\
% MultipleIntermediate &&&&&\\
% computeProduct* & 0.5 & 1 & 0.5 & 0 & \textbf{2} \\
% multiply* & 0.5 & 1 & 0.5 & 0 & \textbf{2} \\
% greatestCommonDivisor & 0 & 0 & 0 & 1 & \textbf{1} \\
% isMultipleOf & 0 & 0 & 0 & 0 & \textbf{0} \\
% lowestCommonMultiple & 0 & 0 & 0 & 0 & \textbf{0} \\
% multiplyAndAdd & 0 & 0 & 0 & 0 & \textbf{0} \\
% \hline
% \textit{H} & & & & &\\
% computeLcmFromGcd & 1 & 1 & 1 & 0.5 & \textbf{3.5} \\
% divide* & 0.5 & 1 & 0.5 & 0.5 & \textbf{2.5} \\
% lcmCalculation & 1 & 0.5 & 0.5 & 0.5 & \textbf{2.5} \\
% divideValues* & 0.5 & 1 & 0.5 & 0 & \textbf{2} \\
% leastCommonMultiple & 0 & 0 & 0 & 1 & \textbf{1} \\
% highestCommonDivisor & 0 & 0 & 0 & 0 & \textbf{0} \\
% divideAndSubtract & 0 & 0 & 0 & 0 & \textbf{0} \\
% \bottomrule
% \end{tabular}
% \end{table}

\begin{table}[h]
\caption{\review{LLM judgment results for Case Study \ref{cs:reachable} (\textit{reachability}). Judges assign a score of 1 for correct and precise names, 0.5 for imprecise or overly general names that are still correct, and 0 for incorrect names. The final score, shown in the last column, is the average of the individual scores given by the judges. Names that can be considered correct are marked with an asterisk (``*'').}}
\label{tab:reachable_jdg_results}
\centering
\begin{tabular}{@{}lcccc@{}}
\toprule
Predicate & ChatGPT-5\footnotemark[1] & Gemini & Command R+ & \textbf{\textbf{Score}} \\
\midrule
\textit{inv1} & & & &\\
directlyConnected* & 1 & 1 & 0,5 & \textbf{0.833}\\
directConnection* & 1 & 1 & 0,5 & \textbf{0.833}\\
isConnected* & 1 & 0,5 & 1 & \textbf{0.833}\\
can reach & 0 & 0,5 & 0,5 & \textbf{0.333}\\
\bottomrule
\end{tabular}
\footnotetext[1]{Since ChatGPT-4o and ChatGPT-o3mini were no longer available at the time of the experiment, ChatGPT-5 was used instead.}
\end{table}

\review{
\subsection{Human judgment results}
Although all the proposed alternatives in the \textit{coauthors} case study 
%can be considered correct, 
are plausible, human judges did not select the most straightforward option for predicate P, that is, \texttt{coauthors}. This behavior aligns with that observed with the \glspl{llm}. 
They also correctly identified the appropriate names for predicates h0 and h3 in the \textit{grandparent} and \textit{cousins} case studies, respectively.
In the \textit{lcm} case study, human judges successfully selected the correct names for predicates G and H, whereas the \glspl{llm} judges failed to do so.
Averaged scores assigned by human judges for these case studies are summarized in Table~\ref{tab:body_hj}.
}

\review{
Table~\ref{tab:family_hj} presents the results of the human evaluation for the proposed names in the \textit{family} case study. For each predicate, the correct name received the highest overall score. However, it is noteworthy that several human judges assigned non-zero scores to incorrect names. For instance, in the case of predicate h0, 4 out of 14 judges assigned a score of 1, and another 4 assigned a score of 0.5 to the incorrect name \texttt{ancestor}.
}

\review{
For the \textit{math} case study, human judges consistently assigned the highest score to the correct name across all predicates, as shown in Table~\ref{tab:math_hj}. Moreover, they successfully identified the correct name for predicate L, that is \texttt{isDivisor}, which none of the \glspl{llm} judges were able to recognize correctly.
}

\begin{table}
\caption{\review{Human judgment results for Case Studies \ref{cs:coauthors} (\textit{coauthors}) and \ref{cs:gp_cs_lcm} (\textit{grandparent}, \textit{cousins}, and \textit{lcm}). Judges assign a score of 1 for correct and precise names, 0.5 for imprecise or overly general names that are still correct, and 0 for incorrect names. The final score, shown in the last column, is the \review{average} of the individual scores given by the judges. Names that can be considered correct are marked with an asterisk (``*'').}}
\label{tab:body_hj}
\centering
% \begin{tabular}{@{}ll@{}}
\begin{tabularx}{0.85\textwidth}{ 
  >{\raggedright\arraybackslash}X
  >{\raggedleft\arraybackslash}X
  }
\toprule
\textbf{Case Study} - \textit{Predicate} & \textbf{Score} \\

\midrule
\textbf{coauthors} - \textit{P} & \\
coauthorsWithCommonPaper & \textbf{0.857}\\
coauthors & \textbf{0.643}\\
coAuthoredPaper & \textbf{0.607}\\
coauthoredResearchPaper & \textbf{0.536}\\
coAuthorResearchers & \textbf{0.464}\\
authoredTogether & \textbf{0.429}\\

\midrule

\textbf{grandparent} - \textit{h0} &\\
parent & \textbf{1.000}\\
renameH0 & \textbf{0.036}\\

\midrule

\textbf{cousins} - \textit{h3} &\\
siblings & \textbf{0.964}\\
sibling & \textbf{0.857}\\
parent & \textbf{0.107}\\
differentParents & \textbf{0.071}\\
isThirdDegreeRelative & \textbf{0.036}\\

\midrule

\textbf{lcm} - \textit{G}\\
computeProduct & \textbf{0.821}\\
multiply & \textbf{0.750}\\
findLeastCommonMultipleIntermediate & \textbf{0.321}\\
isMultipleOf & \textbf{0.214}\\
lowestCommonMultiple & \textbf{0.143}\\
multiplyAndAdd & \textbf{0.143}\\
greatestCommonDivisor & \textbf{0.036}\vspace{0.5em}\\

\textbf{lcm} - \textit{H}\\
divide & \textbf{0.643}\\
divideValues & \textbf{0.643}\\
computeLcmFromGcd & \textbf{0.500}\\
lcmCalculation & \textbf{0.429}\\
leastCommonMultiple & \textbf{0.357}\\
divideAndSubtract & \textbf{0.143}\\
highestCommonDivisor & \textbf{0.143}\\

\bottomrule
\end{tabularx}
\end{table}

\begin{table}[h]
\caption{\review{Human judgment results for Case Study \ref{cs:family} (\textit{family}). Judges assign a score of 1 for correct and precise names, 0.5 for imprecise or overly general names that are still correct, and 0 for incorrect names. The final score, shown in the last column, is the \review{average} of the individual scores given by the judges. Names that can be considered correct are marked with an asterisk (``*'').}}
\label{tab:family_hj}
\centering
\footnotesize  
% \begin{tabular}{@{}ll@{}}
\begin{tabularx}{0.8\textwidth}{ 
  >{\raggedright\arraybackslash}X
  >{\raggedleft\arraybackslash}X
  }
\toprule
\textbf{Predicate} & \textbf{Score}\\
\midrule
\textit{h0} & \\
parent & \textbf{1.000}\\
ancestor & \textbf{0.429}\\
\midrule
\textit{h1} &\\
grandparent & \textbf{1.000}\\
ancestor & \textbf{0.393}\\
\midrule
\textit{h2} &\\
commonAncestor & \textbf{0.929}\\
ancestor & \textbf{0.393}\\
greatAncestor & \textbf{0.250}\\
\midrule
\textit{h3} &\\
sibling & \textbf{1.000}\\
sister & \textbf{0.250}\\
\midrule
\textit{h4} &\\
cousins & \textbf{0.893}\\
cousin & \textbf{0.857}\\
fullSibling & \textbf{0.107}\\
halfSibling & \textbf{0.071}\\
h3 & \textbf{0.071}\\
\bottomrule
\end{tabularx}
\end{table}

\begin{table}[h]
\caption{\review{Human judgment results for Case Study \ref{cs:math} (\textit{math}). Judges assign a score of 1 for correct and precise names, 0.5 for imprecise or overly general names that are still correct, and 0 for incorrect names. The final score, shown in the last column, is the \review{average} of the individual scores given by the judges. Names that can be considered correct are marked with an asterisk (``*'').}}
\label{tab:math_hj}
\centering
\tiny
% \begin{tabular}{@{}ll@{}}
\begin{tabularx}{0.9\textwidth}{ 
  % | >{\raggedright\arraybackslash}X 
  % | >{\centering\arraybackslash}X 
  >{\raggedright\arraybackslash}X
  >{\raggedleft\arraybackslash}X
  }
\toprule
\textbf{Predicate} & \textbf{Score}\\
\midrule
\textit{A} &\\
isNumber & \textbf{0.929}\\
numericValue & \textbf{0.679}\\
isInteger & \textbf{0.250}\\
\midrule
\textit{B} &\\
isEven & \textbf{0.929}\\
even & \textbf{0.750}\\
isOdd & \textbf{0.071}\\
\midrule
\textit{C} &\\
isOdd & \textbf{0.929}\\
odd & \textbf{0.750}\\
isEven & \textbf{0.071}\\
\midrule
\textit{D} &\\
absoluteValue & \textbf{0.964}\\
absValue & \textbf{0.964}\\
negateIfNegative & \textbf{0.250}\\
isPositiveDifference & \textbf{0.000}\\
next & \textbf{0.000}\\
\midrule
\textit{E} &\\
add & \textbf{0.714}\\
sum & \textbf{0.964}\\
\midrule
\textit{F} &\\
subtract & \textbf{0.857}\\
difference & \textbf{0.893}\\
\midrule
\textit{G} &\\
multiply & \textbf{0.786}\\
product & \textbf{0.964}\\
\midrule
\textit{H} &\\
divide & \textbf{0.750}\\
quotient & \textbf{0.679}\\
division & \textbf{0.821}\\
\midrule
\textit{L} &\\
isDivisible & \textbf{0.464}\\
divides & \textbf{0.643}\\
isDivisor & \textbf{0.750}\\
divisibleBy & \textbf{0.393}\\
isDivisibleBy & \textbf{0.500}\\
modulo & \textbf{0.107}\\
\midrule
\textit{M} &\\
gcd & \textbf{0.893}\\
modularArithmetic & \textbf{0.357}\\
customOperation & \textbf{0.286}\\
accumulate & \textbf{0.107}\\
\midrule
\textit{N} &\\
lcm & \textbf{0.893}\\
nestedCalculation & \textbf{0.321}\\
complexOperation & \textbf{0.286}\\
complexCalculation & \textbf{0.286}\\
accumulateSequence & \textbf{0.107}\\
\midrule
\textit{P} &\\
greaterThan & \textbf{1.000}\\
\midrule
\textit{Q} &\\
greaterThanOrEqual & \textbf{0.821}\\
greaterOrEqual & \textbf{0.821}\\
greaterEqual & \textbf{0.500}\\
greaterThanOrEqualTo & \textbf{0.857}\\
\midrule
\textit{R} &\\
lessThan & \textbf{0.929}\\
\midrule
\textit{S} &\\
lessThanOrEqual & \textbf{0.821}\\
lessOrEqual & \textbf{0.821}\\
lessEqual & \textbf{0.500}\\
lessThanOrEqualTo & \textbf{0.857}\\
\midrule
\textit{T} &\\
equal & \textbf{0.750}\\
equalTo & \textbf{0.857}\\
isEqual & \textbf{0.893}\\
\bottomrule
\end{tabularx}
\end{table}

\subsection{Summary of results}
Overall, ChatGPT-4o and ChatGPT-o3mini appear to be the best-performing models, followed by Gemini and Command R+; Llama, FalconMamba, and Falcon3 performed the weakest.

It is worth noting that we chose some relatively small models to test whether they could identify useful names. Llama 3.2 and Falcon3 provided several correct suggestions, though not for every predicate. 
%However, their smaller size means they can be run locally on machines with lower processing power, and fine-tuned through few-shot prompting making this approach accessible to users without high-performance hardware.
%\review{Although few-shot prompting did not lead to improved performance, 
\review{
While few-shot prompting did not enhance performance for Falcon3 and Llama, it led to a substantial improvement for FalconMamba.
Nevertheless, their smaller size allows them to be run locally on machines with lower processing power, making this approach accessible to users without high-performance hardware.
}

The evaluation strategy that we adopted is automatic, meaning that human intervention is not required except for a final assessment, which should be faster than evaluating every single suggestion for each predicate.
It could be argued that an \gls{llm} could make errors not only in making suggestions but also in judging them. Even though in these experiments we asked the \glspl{llm} to score only once, for a more robust evaluation, 
%the process could involve repeating the process multiple times per judge and taking the average. For a more robust evaluation, 
the procedure could be repeated multiple times per judge, and the results averaged.
If a tie occurs between two or more suggestions, a new evaluation could be carried out, or the final decision could be left to domain experts.

\review{
Based on the results of the human evaluation, we can conclude that, overall, \glspl{llm} were generally capable of performing both the renaming and judgment tasks effectively. Their assessments aligned with those of human judges in most cases, with discrepancies emerging primarily in the more challenging or ambiguous predicates.
}

\section{Discussion}\label{sec:disc}
Having unnamed predicates in the output rules may significantly hinder their understanding and reusability, especially for lengthy theories. However, current approaches do not seem to address this issue, with invented predicates typically being left unnamed.
A possible solution is to use placeholders and select names from a predefined pool (\cite{cropper2022ilp30}), but this would require knowing beforehand which predicates should be invented and their arity, which is not always feasible.
While all the predicates could be renamed manually, our approach showed promising results in accomplishing this task automatically.
Even though the involvement of LLMs in logic is still not much investigated, the experiments show that this method was effective in most of the tested cases, even without providing specific contexts or examples. In fact, zero-shot prompting was sufficient, as the \glspl{llm} immediately understood their task. This is particularly convenient when extensive computing power to fine-tune models is not easily accessible, as models for general-purpose tasks, trained on common knowledge data and code can still perform the task (\cite{wu2023exploring}).

% Overall, ChatGPT-4o and ChatGPT-o3mini appear to be the best-performing models, followed by Gemini and Command R+; Llama, FalconMamba, and Falcon3 performed the weakest.
% It is also worth noting that we chose some relatively small models to test whether they could identify useful names. Llama 3.2, FalconMamba, and Falcon3 provided several correct suggestions, though not for every predicate. However, their smaller size means they can be run locally on machines with lower processing power and fine-tuned through few-shot prompting, making this approach accessible to users without high-performance hardware.

% Regarding the evaluation strategy that we adopted, it could be argued that an \gls{llm} could make errors not only in making suggestions but also in scoring them. Even though in these experiments we asked the \glspl{llm} to score only once, for a more robust evaluation, the process could involve repeating the process multiple times per judge and taking the average.
% If a tie occurs between two or more suggestions, a new evaluation could be carried out, or the final decision could be left to domain experts.

\subsection{Limitations}
There are several limitations to this early-stage approach.

The main one is the use of relatively simple examples. The rules express common knowledge and typical relationships, which the models may have already seen in the \gls{lp} codes they have been trained on. Nonetheless, this initial step was necessary to assess whether \glspl{llm} could \review{correctly process} them; otherwise, their ability to grasp even more specific relationships would have seemed unlikely.
Despite the simplicity of the examples, we encountered several challenges throughout the process.
One issue is that, at times, some models fail to follow the request or the output format, which complicates the automation of the entire process. 
For instance, in some preliminary experiments, we noticed that some models also altered the body of the rule,  even if instructed to find a name for the head predicate. This is not desirable, as it could completely change the meaning of the logic theory, which is assumed to be correct. That is why we decided to add “Do NOT change the body of the rules” to the instructions.
Another constraint, ``Give only ONE suggestion for each predicate.”, was added to the prompt to limit the number of suggestions. While one could argue that this might cause the potentially correct answer to be overlooked, we addressed this concern by repeating the request multiple times.
When this behavior persisted, we decided to disregard that particular answer for that specific predicate. Even if one answer is discarded, the presence of multiple models offering suggestions reduces the impact of this issue.
Models also struggled to recognize that some names, despite being written in different formats, were conceptually equivalent (e.g., \textit{lessThan}, \textit{less\_than}, \textit{LessThan}), even when explicitly instructed to treat them as equal. As a result, it was necessary to manually standardize these names, to ensure that they were evaluated as the same. The issue persists to some extent: singular and plural forms are considered different (e.g., \textit{cousins} and \textit{cousin}), as well as names that include a verb or particle (e.g., \textit{isEqual} or \textit{equalTo}).
Lastly, this approach was unsuccessful when tested on predicates from a real-world dataset, Mutagenesis.
It should be noted that nearly all of the predicates' true names were masked, which significantly hindered the models' ability to identify them and grasp the relationship between them. However, such a scenario is highly unlikely, if not entirely unrealistic, since it would mean that the background knowledge is almost empty. In this case, inferring a set of meaningful rules would be highly difficult for any \gls{ml} model.
Nevertheless, our findings hint that domain-specific fine-tuning might be necessary for real-world applications (\cite{rezayi2022agribert}), provided that enough computational resources are available. Alternatively, big models could be replaced by smaller or quantized fine-tuned versions, which are more computationally efficient and have been shown to not significantly decrease the quality of the results (\cite{liu2024understanding,kim2024memory,dettmers2023qlora}). 
%Additionally, few-shot prompting could be applied to further enhance the quality of the answers. %TODO: REF

\section{Conclusion}\label{sec:concl}
In this paper, we have proposed a straightforward pipeline that leverages \glspl{llm} to find suggestions for unnamed predicates in logic theories, since their presence limits the interpretability. Based on the results obtained from some hand-crafted sets of rules, \glspl{llm} appear to be a promising solution to this problem. To the best of our knowledge, this is the first study to address this issue.
The limitations of this study are primarily related to the examples we used, and investigating more complex scenarios is an interesting future work. 
\review{
%An interesting direction for future work involves exploring the extent to which the choice of the logic programming formalism influences the names proposed for invented predicates. While in this work we focused on Prolog, 
We believe that the suitability of the proposed names is primarily determined by the context and semantics of the program and thus developed an approach that is independent of the language. Nevertheless, differences in language-specific conventions may play a role, and investigating this aspect could offer further insights into the generalization ability of LLM-based renaming approaches.
}
Another topic worth exploring is the use of fine-tuned \glspl{llm} for domain-specific naming suggestions, as well as the development of quantized versions that can run locally on machines with lower processing power.
Future research could also consider the potential of \glspl{llm} for directly performing PI and their application to other approaches such as \gls{lfit} (\cite{ribeiro2021lfit}) to reduce the number of rules.

\backmatter

% \section*{Abbreviations}
% \glsaddall
% \printglossary[title={},type=\acronymtype]

\section*{Declarations}
% Some journals require declarations to be submitted in a standardised format. Please check the Instructions for Authors of the journal to which you are submitting to see if you need to complete this section. If yes, your manuscript must contain the following sections under the heading `Declarations':

% \begin{itemize}
% \item Funding
% \item Conflict of interest/Competing interests (check journal-specific guidelines for which heading to use)
% \item Ethics approval and consent to participate
% \item Consent for publication
% \item Data availability 
% \item Materials availability
% \item Code availability 
% \item Author contribution
% \end{itemize}

% \noindent
% If any of the sections are not relevant to your manuscript, please include the heading and write `Not applicable' for that section. 

\subsection*{Funding}
Elisabetta Gentili contributed to this publication while attending the PhD programme in Engineering Science at the University of Ferrara, Cycle XXXVIII, with the support of a scholarship financed by the Ministerial Decree no. 351 of 9th April 2022, based on the NRRP - funded by the European Union - NextGenerationEU - Mission 4 ``Education and Research'', Component 1 ``Enhancement of the offer of educational services: from nurseries to universities'' - Investment 4.1 ``Extension of the number of research doctorates and innovative doctorates for public administration and cultural heritage''.

This work has been partially supported by Spoke 1 ``FutureHPC \& BigData'' of the Italian Research Center on High-Performance Computing, Big Data and Quantum Computing (ICSC) funded by MUR Missione 4 - Next Generation EU (NGEU) and by Partenariato Esteso PE00000013 - ``FAIR - Future Artificial Intelligence Research'' - Spoke 8 ``Pervasive AI'', funded by MUR through PNRR - M4C2 - Investimento 1.3 (Decreto Direttoriale MUR n. 341 of 15th March 2022) under the Next Generation EU (NGEU).

Elisabetta Gentili and Fabrizio Riguzzi are members of the Gruppo Nazionale Calcolo Scientifico -- Istituto Nazionale di Alta Matematica (GNCS-INdAM).

Katsumi Inoue has been supported in part by JST CREST Grant Number
JPMJCR22D3, Japan.

Tony Ribeiro has been funded by the European Union. Views
and opinions expressed are however those of the author(s) only and do not necessarily reflect those of the European Union or ERCEA. Neither the European Union nor the granting authority can be held responsible for them.

\subsection*{Competing interests}
The authors declare none.

% \begin{appendices}

% \section{Section title of first appendix}\label{secA1}

% An appendix contains supplementary information that is not an essential part of the text itself but which may be helpful in providing a more comprehensive understanding of the research problem or it is information that is too cumbersome to be included in the body of the paper.

%%=============================================%%
%% For submissions to Nature Portfolio Journals %%
%% please use the heading ``Extended Data''.   %%
%%=============================================%%

%%=============================================================%%
%% Sample for another appendix section			       %%
%%=============================================================%%

%% \section{Example of another appendix section}\label{secA2}%
%% Appendices may be used for helpful, supporting or essential material that would otherwise 
%% clutter, break up or be distracting to the text. Appendices can consist of sections, figures, 
%% tables and equations etc.

% \end{appendices}

%%===========================================================================================%%
%% If you are submitting to one of the Nature Portfolio journals, using the eJP submission   %%
%% system, please include the references within the manuscript file itself. You may do this  %%
%% by copying the reference list from your .bbl file, paste it into the main manuscript .tex %%
%% file, and delete the associated \verb+\bibliography+ commands.                            %%
%%=====

\bibliography{sn-bibliography}% common bib file

\begin{thebibliography}{93}
\providecommand{\natexlab}[1]{#1}
\providecommand{\url}[1]{{#1}}
\providecommand{\urlprefix}{URL }
\providecommand{\doi}[1]{\url{https://doi.org/#1}}
\providecommand{\eprint}[2][]{\url{#2}}
 \bibcommenthead

\bibitem[{Abiteboul et~al.(1995)Abiteboul, Hull, and Vianu}]{abiteboul1995foundations}
Abiteboul S, Hull R, Vianu V (1995) Foundations of Databases. Addison-Wesley, Reading, MA, USA

\bibitem[{Alammar and Grootendorst(2024)}]{alammar2024hands}
Alammar J, Grootendorst M (2024) Hands-On Large Language Models. O'Reilly, Sebastopol, CA, USA, \urlprefix\url{https://www.oreilly.com/library/view/hands-on-large-language/9781098150952/}

\bibitem[{Basu et~al.(2021)Basu, Murugesan, Atzeni, Kapanipathi, Talamadupula, Klinger, Campbell, Sachan, and Gupta}]{basu2021hybrid}
Basu K, Murugesan K, Atzeni M, et~al (2021) A hybrid neuro-symbolic approach for text-based games using inductive logic programming. In: Combining learning and reasoning: programming languages, formalisms, and representations

\bibitem[{Bommasani et~al.(2021)Bommasani, Hudson, Adeli, Altman, Arora, von Arx, Bernstein, Bohg, Bosselut, Brunskill et~al.}]{bommasani2021opportunities}
Bommasani R, Hudson DA, Adeli E, et~al (2021) On the opportunities and risks of foundation models. arXiv preprint arXiv:210807258

\bibitem[{Bordes et~al.(2013)Bordes, Usunier, Garcia-Duran, Weston, and Yakhnenko}]{bordes2013translating}
Bordes A, Usunier N, Garcia-Duran A, et~al (2013) Translating embeddings for modeling multi-relational data. Advances in neural information processing systems 26

\bibitem[{Brown et~al.(2020)Brown, Mann, Ryder, Subbiah, Kaplan, Dhariwal, Neelakantan, Shyam, Sastry, Askell, Agarwal, Herbert-Voss, Krueger, Henighan, Child, Ramesh, Ziegler, Wu, Winter, Hesse, Chen, Sigler, Litwin, Gray, Chess, Clark, Berner, McCandlish, Radford, Sutskever, and Amodei}]{brown2020language}
Brown T, Mann B, Ryder N, et~al (2020) {Language Models are Few-Shot Learners}. In: Larochelle H, Ranzato M, Hadsell R, et~al (eds) Advances in Neural Information Processing Systems, vol~33. Curran Associates, Inc., Red Hook, NY, USA, p 1877--1901, \urlprefix\url{https://proceedings.neurips.cc/paper\_files/paper/2020/file/1457c0d6bfcb4967418bfb8ac142f64a-Paper.pdf}

\bibitem[{Chen et~al.(2021)Chen, Tworek, Jun, Yuan, Pinto, Kaplan, Edwards, Burda, Joseph, Brockman et~al.}]{chen2021evaluating}
Chen M, Tworek J, Jun H, et~al (2021) Evaluating large language models trained on code. arXiv preprint arXiv:210703374

\bibitem[{Cobbe et~al.(2021)Cobbe, Kosaraju, Bavarian, Chen, Jun, Kaiser, Plappert, Tworek, Hilton, Nakano et~al.}]{cobbe2021training}
Cobbe K, Kosaraju V, Bavarian M, et~al (2021) Training verifiers to solve math word problems. arXiv preprint arXiv:211014168

\bibitem[{{Cohere For AI}(2024)}]{cohere2024}
{Cohere For AI} (2024) c4ai-command-r-plus-08-2024. \doi{10.57967/hf/3135}, \urlprefix\url{https://huggingface.co/CohereForAI/c4ai-command-r-plus-08-2024}

\bibitem[{Colmerauer and Roussel(1996)}]{colmerauer1996birth}
Colmerauer A, Roussel P (1996) The birth of Prolog, Association for Computing Machinery, New York, NY, USA, p 331–367. \urlprefix\url{https://doi.org/10.1145/234286.1057820}

\bibitem[{Creswell et~al.(2022)Creswell, Shanahan, and Higgins}]{creswell2022selection}
Creswell A, Shanahan M, Higgins I (2022) Selection-inference: Exploiting large language models for interpretable logical reasoning. arXiv preprint arXiv:220509712

\bibitem[{Cropper and Duman{\v{c}}i{\'c}(2022)}]{cropper2022ilp30}
Cropper A, Duman{\v{c}}i{\'c} S (2022) Inductive logic programming at 30: a new introduction. Journal of Artificial Intelligence Research 74:765--850

\bibitem[{Cropper and Morel(2021{\natexlab{a}})}]{cropper2021learning}
Cropper A, Morel R (2021{\natexlab{a}}) Learning programs by learning from failures. Machine Learning 110(4):801--856

\bibitem[{Cropper and Morel(2021{\natexlab{b}})}]{cropper2021predicate}
Cropper A, Morel R (2021{\natexlab{b}}) Predicate invention by learning from failures. arXiv preprint arXiv:210414426

\bibitem[{Cunnington et~al.(2024)Cunnington, Law, Lobo, and Russo}]{cunnington2024role}
Cunnington D, Law M, Lobo J, et~al (2024) The role of foundation models in neuro-symbolic learning and reasoning. In: Besold TR, d'Avila Garcez A, Jimenez-Ruiz E, et~al (eds) Neural-Symbolic Learning and Reasoning. Springer Nature Switzerland, Cham, pp 84--100

\bibitem[{Dash et~al.(2024)Dash, Mathur, and Patil}]{dash2024enhancing}
Dash A, Mathur B, Patil P (2024) Enhancing carbon emission tracking with gemini: An ai solution for sustainable development. \doi{10.13140/RG.2.2.26165.90089}

\bibitem[{De~Giorgis et~al.(2024)De~Giorgis, Gangemi, and Russo}]{de2024neurosymbolic}
De~Giorgis S, Gangemi A, Russo A (2024) Neurosymbolic graph enrichment for grounded world models. arXiv preprint arXiv:241112671

\bibitem[{DeepMind(2024)}]{gemini2024}
DeepMind G (2024) {Gemini 1.5 Flash}. \urlprefix\url{https://ai.google.dev/gemini-api/docs/models/gemini?hl=it#gemini-1.5-flash}

\bibitem[{Dettmers et~al.(2023)Dettmers, Pagnoni, Holtzman, and Zettlemoyer}]{dettmers2023qlora}
Dettmers T, Pagnoni A, Holtzman A, et~al (2023) Qlora: Efficient finetuning of quantized llms. In: Oh A, Naumann T, Globerson A, et~al (eds) Advances in Neural Information Processing Systems, vol~36. Curran Associates, Inc., Red Hook, NY, USA, pp 10088--10115

\bibitem[{Fadja et~al.(2017)Fadja, Lamma, and Riguzzi}]{fadja2017deep}
Fadja AN, Lamma E, Riguzzi F (2017) Deep probabilistic logic programming. In: Plp@ Ilp, pp 3--14

\bibitem[{{Falcon-LLM Team}(2024)}]{falcon3}
{Falcon-LLM Team} (2024) {The Falcon 3 Family of Open Models}. \urlprefix\url{https://huggingface.co/blog/falcon3}

\bibitem[{Feuerriegel et~al.(2024)Feuerriegel, Hartmann, Janiesch, and Zschech}]{feuerriegel2024generative}
Feuerriegel S, Hartmann J, Janiesch C, et~al (2024) Generative ai. Business \& Information Systems Engineering 66(1):111--126

\bibitem[{Gandarela et~al.(2024)Gandarela, Carvalho, and Freitas}]{gandarela2024inductive}
Gandarela JP, Carvalho DS, Freitas A (2024) Inductive learning of logical theories with {LLMs}: A complexity-graded analysis. arXiv preprint arXiv:240816779

\bibitem[{Gelfond and Lifschitz(1988)}]{gelfond1988stablemodelsem}
Gelfond M, Lifschitz V (1988) The stable model semantics for logic programming. In: ICLP/SLP, \urlprefix\url{https://api.semanticscholar.org/CorpusID:261517573}

\bibitem[{Gu and Dao(2023)}]{gu2023mamba}
Gu A, Dao T (2023) Mamba: Linear-time sequence modeling with selective state spaces. arXiv preprint arXiv:231200752

\bibitem[{Gulwani(2011)}]{gulwani2011automating}
Gulwani S (2011) Automating string processing in spreadsheets using input-output examples. ACM Sigplan Notices 46(1):317--330

\bibitem[{Han et~al.(2024)Han, Ransom, Perfors, and Kemp}]{han2024inductive}
Han SJ, Ransom KJ, Perfors A, et~al (2024) Inductive reasoning in humans and large language models. Cognitive Systems Research 83:101155. \doi{https://doi.org/10.1016/j.cogsys.2023.101155}, \urlprefix\url{https://www.sciencedirect.com/science/article/pii/S1389041723000839}

\bibitem[{Hinton(2015)}]{hinton2015distilling}
Hinton G (2015) Distilling the knowledge in a neural network. arXiv preprint arXiv:150302531

\bibitem[{Hou et~al.(2024)Hou, Zhao, Liu, Yang, Wang, Li, Luo, Lo, Grundy, and Wang}]{hou2024se}
Hou X, Zhao Y, Liu Y, et~al (2024) Large language models for software engineering: A systematic literature review. ACM Transactions on Software Engineering and Methodology 33(8):1--79

\bibitem[{Inoue et~al.(2010)Inoue, Furukawa, Kobayashi, and Nabeshima}]{inoue2010discovering}
Inoue K, Furukawa K, Kobayashi I, et~al (2010) Discovering rules by meta-level abduction. In: Inductive Logic Programming: 19th International Conference, ILP 2009, Leuven, Belgium, July 02-04, 2009. Revised Papers 19, Springer, pp 49--64

\bibitem[{Inoue et~al.(2013)Inoue, Doncescu, and Nabeshima}]{inoue2013completing}
Inoue K, Doncescu A, Nabeshima H (2013) Completing causal networks by meta-level abduction. Machine learning 91(2):239--277

\bibitem[{Inoue et~al.(2014)Inoue, Ribeiro, and Sakama}]{inoue2014learning}
Inoue K, Ribeiro T, Sakama C (2014) Learning from interpretation transition. Machine Learning 94:51--79

\bibitem[{Ishay et~al.(2023)Ishay, Yang, and Lee}]{ishay2023leveraging}
Ishay A, Yang Z, Lee J (2023) Leveraging large language models to generate answer set programs. arXiv preprint arXiv:230707699

\bibitem[{Jeblick et~al.(2024)Jeblick, Schachtner, Dexl, Mittermeier, St{\"u}ber, Topalis, Weber, Wesp, Sabel, Ricke et~al.}]{jeblick2024chatgpt}
Jeblick K, Schachtner B, Dexl J, et~al (2024) Chatgpt makes medicine easy to swallow: an exploratory case study on simplified radiology reports. European radiology 34(5):2817--2825

\bibitem[{Jiang et~al.(2024)Jiang, Wang, Shen, Kim, and Kim}]{jiang2024codegen}
Jiang J, Wang F, Shen J, et~al (2024) A survey on large language models for code generation. arXiv preprint arXiv:240600515

\bibitem[{Kaddour et~al.(2023)Kaddour, Harris, Mozes, Bradley, Raileanu, and McHardy}]{kaddour2023challenges}
Kaddour J, Harris J, Mozes M, et~al (2023) Challenges and applications of large language models. arXiv preprint arXiv:230710169

\bibitem[{Kakas et~al.(1992)Kakas, Kowalski, and Toni}]{kakas1992abductive}
Kakas AC, Kowalski RA, Toni F (1992) Abductive logic programming. Journal of logic and computation 2(6):719--770

\bibitem[{Kareem et~al.(2025)Kareem, Gallagher, Borroto, Ricca, and Russo}]{kareem2025using}
Kareem I, Gallagher K, Borroto M, et~al (2025) Using learning from answer sets for robust question answering with llm. In: Dodaro C, Gupta G, Martinez MV (eds) Logic Programming and Nonmonotonic Reasoning. Springer Nature Switzerland, Cham, pp 112--125

\bibitem[{Kim et~al.(2023)Kim, Lee, Kim, Park, Yoo, Kwon, and Lee}]{kim2024memory}
Kim J, Lee JH, Kim S, et~al (2023) Memory-efficient fine-tuning of compressed large language models via sub-4-bit integer quantization. In: Oh A, Naumann T, Globerson A, et~al (eds) Advances in Neural Information Processing Systems, vol~36. Curran Associates, Inc., Red Hook, NY, USA, pp 36187--36207

\bibitem[{Kobayashi and Furukawa(2009)}]{kobayashi2009hypothesis}
Kobayashi I, Furukawa K (2009) Hypothesis selection using domain theory in rule abductive support for skills. In: SIG-SKL (Skill Science). Japanese Society for Artificial Intelligence (in Japanese)

\bibitem[{Kojima et~al.(2022)Kojima, Gu, Reid, Matsuo, and Iwasawa}]{kojima2022advances}
Kojima T, Gu SS, Reid M, et~al (2022) Large language models are zero-shot reasoners. In: Koyejo S, Mohamed S, Agarwal A, et~al (eds) Advances in Neural Information Processing Systems, vol~35. Curran Associates, Inc., Red Hook, NY, USA, pp 22199--22213, \urlprefix\url{https://proceedings.neurips.cc/paper\_files/paper/2022/file/8bb0d291acd4acf06ef112099c16f326-Paper-Conference.pdf}

\bibitem[{Kok and Domingos(2005)}]{kok2005learning}
Kok S, Domingos P (2005) Learning the structure of markov logic networks. In: Proceedings of the 22nd international conference on Machine learning, pp 441--448

\bibitem[{Koubaa et~al.(2023)Koubaa, Boulila, Ghouti, Alzahem, and Latif}]{koubaa2023exploring}
Koubaa A, Boulila W, Ghouti L, et~al (2023) Exploring chatgpt capabilities and limitations: a survey. IEEE Access

\bibitem[{Kowalski(1988)}]{kowalski1988logicprog}
Kowalski RA (1988) The early years of logic programming. Commun ACM 31:38--43. \urlprefix\url{https://api.semanticscholar.org/CorpusID:12259230}

\bibitem[{Kramer(1995)}]{kramer1995predicate}
Kramer S (1995) Predicate invention: A comprehensive view. Rapport technique OFAI-TR-95-32, Austrian Research Institute for Artificial Intelligence, Vienna

\bibitem[{Law et~al.(2014)Law, Russo, and Broda}]{law2014inductive}
Law M, Russo A, Broda K (2014) Inductive learning of answer set programs. In: Logics in Artificial Intelligence: 14th European Conference, JELIA 2014, Funchal, Madeira, Portugal, September 24-26, 2014. Proceedings 14, Springer, pp 311--325

\bibitem[{Law et~al.(2020)Law, Russo, and Broda}]{law2020ilasp}
Law M, Russo A, Broda K (2020) The ilasp system for inductive learning of answer set programs. arXiv preprint arXiv:200500904

\bibitem[{Li et~al.(2024)Li, Cao, Xu, Jiang, Liu, Teo, Lin, and Liu}]{li2024llms}
Li Z, Cao Y, Xu X, et~al (2024) Llms for relational reasoning: How far are we? In: Proceedings of the 1st International Workshop on Large Language Models for Code, pp 119--126

\bibitem[{Liu et~al.(2024{\natexlab{a}})Liu, Lin, Hewitt, Paranjape, Bevilacqua, Petroni, and Liang}]{liu2024lost}
Liu NF, Lin K, Hewitt J, et~al (2024{\natexlab{a}}) Lost in the middle: How language models use long contexts. Transactions of the Association for Computational Linguistics 12:157--173

\bibitem[{Liu et~al.(2024{\natexlab{b}})Liu, He, Han, Zhang, Liu, Tian, Zhang, Wang, Gao, Zhong et~al.}]{liu2024understanding}
Liu Y, He H, Han T, et~al (2024{\natexlab{b}}) Understanding llms: A comprehensive overview from training to inference. arXiv preprint arXiv:240102038

\bibitem[{Magdy et~al.(2024)Magdy, Alwajih, Kwon, Abdel-Salam, and Abdul-Mageed}]{magdy2024gazelle}
Magdy SM, Alwajih F, Kwon SY, et~al (2024) Gazelle: An instruction dataset for arabic writing assistance. arXiv preprint arXiv:241018163

\bibitem[{Meta(2024{\natexlab{a}})}]{llama32mc}
Meta (2024{\natexlab{a}}) {Llama 3.2 Model Card}. \urlprefix\url{https://github.com/meta-llama/llama-models/blob/main/models/llama3_2/MODEL_CARD.md}

\bibitem[{Meta(2024{\natexlab{b}})}]{llama32intro}
Meta (2024{\natexlab{b}}) {Llama 3.2: Revolutionizing edge AI and vision with open, customizable models}. \urlprefix\url{https://github.com/meta-llama/llama-models/blob/main/models/llama3_2/MODEL_CARD.md}

\bibitem[{Min et~al.(2022)Min, Lyu, Holtzman, Artetxe, Lewis, Hajishirzi, and Zettlemoyer}]{min2022rethinking}
Min S, Lyu X, Holtzman A, et~al (2022) { Rethinking the Role of Demonstrations: What makes In-context Learning Work? }. In: EMNLP

\bibitem[{Mondillo et~al.(2024)Mondillo, Frattolillo, Colosimo, Perrotta, Di~Sessa, Guarino, Miraglia~del Giudice, and Marzuillo}]{mondillo2024basal}
Mondillo G, Frattolillo V, Colosimo S, et~al (2024) Basal knowledge in the field of pediatric nephrology and its enhancement following specific training of chatgpt-4 “omni” and gemini 1.5 flash. Pediatric Nephrology pp 1--7

\bibitem[{Mooney(1996)}]{mooney1996inductive}
Mooney RJ (1996) Inductive logic programming for natural language processing. In: International conference on inductive logic programming, Springer, pp 1--22

\bibitem[{{Moradi Dakhel} et~al.(2023){Moradi Dakhel}, Majdinasab, Nikanjam, Khomh, Desmarais, and Jiang}]{moradi2023copilot}
{Moradi Dakhel} A, Majdinasab V, Nikanjam A, et~al (2023) Github copilot ai pair programmer: Asset or liability? Journal of Systems and Software 203:111734. \doi{https://doi.org/10.1016/j.jss.2023.111734}, \urlprefix\url{https://www.sciencedirect.com/science/article/pii/S0164121223001292}

\bibitem[{Muggleton(1991)}]{muggleton1991inductive}
Muggleton S (1991) Inductive logic programming. New generation computing 8:295--318

\bibitem[{Muggleton and Buntine(1988)}]{muggleton1988machine}
Muggleton S, Buntine W (1988) Machine invention of first-order predicates by inverting resolution. In: Machine Learning Proceedings 1988. Elsevier, San Francisco (CA), p 339--352

\bibitem[{Muggleton et~al.(2012)Muggleton, De~Raedt, Poole, Bratko, Flach, Inoue, and Srinivasan}]{muggleton2012ilp}
Muggleton S, De~Raedt L, Poole D, et~al (2012) Ilp turns 20: biography and future challenges. Machine learning 86:3--23

\bibitem[{Muggleton et~al.(2015)Muggleton, Lin, and Tamaddoni-Nezhad}]{muggleton2015meta}
Muggleton SH, Lin D, Tamaddoni-Nezhad A (2015) Meta-interpretive learning of higher-order dyadic datalog: Predicate invention revisited. Machine Learning 100(1):49--73

\bibitem[{Nguembang~Fadja et~al.(2021)Nguembang~Fadja, Riguzzi, and Lamma}]{nguembang2021learning}
Nguembang~Fadja A, Riguzzi F, Lamma E (2021) Learning hierarchical probabilistic logic programs. Machine Learning 110(7):1637--1693

\bibitem[{OpenAI(2024{\natexlab{a}})}]{openai2024gpt4ocard}
OpenAI (2024{\natexlab{a}}) {GPT-4o System Card}. \urlprefix\url{https://openai.com/index/gpt-4o-system-card}

\bibitem[{OpenAI(2024{\natexlab{b}})}]{openai2024hellogpt4o}
OpenAI (2024{\natexlab{b}}) {Hello GPT-4o}. \urlprefix\url{https://openai.com/index/hello-gpt-4o}

\bibitem[{OpenAI(2025)}]{openai2025gpto3minicard}
OpenAI (2025) {OpenAI o3-mini System Card}. \urlprefix\url{https://openai.com/index/o3-mini-system-card}

\bibitem[{Pearce et~al.(2023)Pearce, Tan, Ahmad, Karri, and Dolan-Gavitt}]{pearce2023examining}
Pearce H, Tan B, Ahmad B, et~al (2023) Examining zero-shot vulnerability repair with large language models. In: 2023 IEEE Symposium on Security and Privacy (SP), pp 2339--2356, \doi{10.1109/SP46215.2023.10179324}

\bibitem[{Raffel et~al.(2020)Raffel, Shazeer, Roberts, Lee, Narang, Matena, Zhou, Li, and Liu}]{raffel2020exploring}
Raffel C, Shazeer N, Roberts A, et~al (2020) Exploring the limits of transfer learning with a unified text-to-text transformer. Journal of Machine Learning Research 21(140):1--67. \urlprefix\url{http://jmlr.org/papers/v21/20-074.html}

\bibitem[{Rao et~al.(2023)Rao, Pang, Kim, Kamineni, Lie, Prasad, Landman, Dreyer, and Succi}]{rao2023assessing}
Rao A, Pang M, Kim J, et~al (2023) Assessing the utility of chatgpt throughout the entire clinical workflow: development and usability study. Journal of Medical Internet Research 25:e48659

\bibitem[{Rezayi et~al.(2022)Rezayi, Liu, Wu, Dhakal, Ge, Zhen, Liu, and Li}]{rezayi2022agribert}
Rezayi S, Liu Z, Wu Z, et~al (2022) Agribert: Knowledge-infused agricultural language models for matching food and nutrition. In: IJCAI, pp 5150--5156

\bibitem[{Ribeiro et~al.(2021)Ribeiro, Folschette, Magnin, and Inoue}]{ribeiro2021lfit}
Ribeiro T, Folschette M, Magnin M, et~al (2021) {Learning any memory-less discrete semantics for dynamical systems represented by logic programs}. {Machine Learning} \doi{10.1007/s10994-021-06105-4}, \urlprefix\url{https://hal.science/hal-02925942}

\bibitem[{Riguzzi(2023)}]{riguzzi2023foundations}
Riguzzi F (2023) Foundations of Probabilistic Logic Programming: Languages, Semantics, Inference and Learning (2nd ed.). River Publishers, Gistrup, Denmark

\bibitem[{Sato et~al.(2018)Sato, Inoue, and Sakama}]{sato2018abducing}
Sato T, Inoue K, Sakama C (2018) Abducing relations in continuous spaces. In: IJCAI, pp 1956--1962

\bibitem[{Sha et~al.(2024)Sha, Shindo, Kersting, and Dhami}]{sha2024neuro}
Sha J, Shindo H, Kersting K, et~al (2024) Neuro-symbolic predicate invention: Learning relational concepts from visual scenes. Neurosymbolic Artificial Intelligence (Preprint):1--26

\bibitem[{Srinivasan et~al.(1996)Srinivasan, Muggleton, Sternberg, and King}]{srinivasan1996theories}
Srinivasan A, Muggleton SH, Sternberg MJ, et~al (1996) Theories for mutagenicity: A study in first-order and feature-based induction. Artificial Intelligence 85(1-2):277--299

\bibitem[{Srinivasan et~al.(1997)Srinivasan, King, Muggleton, and Sternberg}]{srinivasan1997carcinogenesis}
Srinivasan A, King RD, Muggleton SH, et~al (1997) Carcinogenesis predictions using ilp. In: International Conference on Inductive Logic Programming, Springer, pp 273--287

\bibitem[{Srivastava et~al.(2022)Srivastava, Rastogi, Rao, Shoeb, Abid, Fisch, Brown, Santoro, Gupta, Garriga-Alonso et~al.}]{srivastava2022beyond}
Srivastava A, Rastogi A, Rao A, et~al (2022) Beyond the imitation game: Quantifying and extrapolating the capabilities of language models. arXiv preprint arXiv:220604615

\bibitem[{Stahl(1993)}]{stahl1993predicate}
Stahl I (1993) Predicate invention in ilp—an overview. In: European conference on machine learning, Springer, pp 311--322

\bibitem[{Stahl(1995)}]{stahl1995appropriateness}
Stahl I (1995) The appropriateness of predicate invention as bias shift operation in ilp. Machine Learning 20:95--117

\bibitem[{Talmor et~al.(2018)Talmor, Herzig, Lourie, and Berant}]{talmor2018commonsenseqa}
Talmor A, Herzig J, Lourie N, et~al (2018) Commonsenseqa: A question answering challenge targeting commonsense knowledge. arXiv preprint arXiv:181100937

\bibitem[{Tarau(2025)}]{tarau2025llm}
Tarau P (2025) On llm-generated logic programs and their inference execution methods. arXiv preprint arXiv:250209209

\bibitem[{Touvron et~al.(2023)Touvron, Lavril, Izacard, Martinet, Lachaux, Lacroix, Rozi{\`e}re, Goyal, Hambro, Azhar et~al.}]{touvron2023llama}
Touvron H, Lavril T, Izacard G, et~al (2023) Llama: Open and efficient foundation language models. arXiv preprint arXiv:230213971

\bibitem[{Valmeekam et~al.(2022)Valmeekam, Olmo, Sreedharan, and Kambhampati}]{valmeekam2022large}
Valmeekam K, Olmo A, Sreedharan S, et~al (2022) Large language models still can't plan (a benchmark for llms on planning and reasoning about change). In: NeurIPS 2022 Foundation Models for Decision Making Workshop

\bibitem[{Vaswani et~al.(2017)Vaswani, Shazeer, Parmar, Uszkoreit, Jones, Gomez, Kaiser, and Polosukhin}]{NIPS20173f5ee243}
Vaswani A, Shazeer N, Parmar N, et~al (2017) Attention is all you need. In: Guyon I, Luxburg UV, Bengio S, et~al (eds) Advances in Neural Information Processing Systems, vol~30. Curran Associates, Inc., Red Hook, NY, USA, \urlprefix\url{https://proceedings.neurips.cc/paper\_files/paper/2017/file/3f5ee243547dee91fbd053c1c4a845aa-Paper.pdf}

\bibitem[{Vennekens et~al.(2004)Vennekens, Verbaeten, and Bruynooghe}]{vennekens2004logic}
Vennekens J, Verbaeten S, Bruynooghe M (2004) Logic programs with annotated disjunctions. In: Logic Programming: 20th International Conference, ICLP 2004, Saint-Malo, France, September 6-10, 2004. Proceedings 20, Springer, pp 431--445

\bibitem[{Wang et~al.(2023)Wang, Zelikman, Poesia, Pu, Haber, and Goodman}]{wang2023hypothesis}
Wang R, Zelikman E, Poesia G, et~al (2023) Hypothesis search: Inductive reasoning with language models. arXiv preprint arXiv:230905660

\bibitem[{Wei et~al.(2022)Wei, Wang, Schuurmans, Bosma, Xia, Chi, Le, Zhou et~al.}]{wei2022chain}
Wei J, Wang X, Schuurmans D, et~al (2022) Chain-of-thought prompting elicits reasoning in large language models. Advances in neural information processing systems 35:24824--24837

\bibitem[{Wolf et~al.(2020)Wolf, Debut, Sanh, Chaumond, Delangue, Moi, Cistac, Rault, Louf, Funtowicz, Davison, Shleifer, von Platen, Ma, Jernite, Plu, Xu, Scao, Gugger, Drame, Lhoest, and Rush}]{wolf2020transformers}
Wolf T, Debut L, Sanh V, et~al (2020) Transformers: State-of-the-art natural language processing. In: Proceedings of the 2020 Conference on Empirical Methods in Natural Language Processing: System Demonstrations. Association for Computational Linguistics, Online, pp 38--45, \urlprefix\url{https://www.aclweb.org/anthology/2020.emnlp-demos.6}

\bibitem[{Wu et~al.(2025)Wu, Zhang, Cao, Yu, Liu, Zhao, Li, Dai, Ma, Li, Liu, Li, Shen, Li, Zhu, and Liu}]{wu2023exploring}
Wu Z, Zhang L, Cao C, et~al (2025) Exploring the trade-offs: Unified large language models vs local fine-tuned models for highly-specific radiology nli task. IEEE Transactions on Big Data 11(3):1027--1041. \doi{10.1109/TBDATA.2025.3536928}

\bibitem[{Yang et~al.(2020)Yang, Wang, Shi, Tadepalli, Lee, and Tu}]{yang2020decoder}
Yang Y, Wang L, Shi S, et~al (2020) On the sub-layer functionalities of transformer decoder. In: Cohn T, He Y, Liu Y (eds) Findings of the Association for Computational Linguistics: EMNLP 2020. Association for Computational Linguistics, Online, pp 4799--4811, \doi{10.18653/v1/2020.findings-emnlp.432}, \urlprefix\url{https://aclanthology.org/2020.findings-emnlp.432/}

\bibitem[{Yang et~al.(2024)Yang, Dong, Du, Cheng, Cambria, Liu, Gao, and Wei}]{yang2024language}
Yang Z, Dong L, Du X, et~al (2024) Language models as inductive reasoners. In: Graham Y, Purver M (eds) Proceedings of the 18th Conference of the European Chapter of the Association for Computational Linguistics (Volume 1: Long Papers). Association for Computational Linguistics, St. Julian{'}s, Malta, pp 209--225, \urlprefix\url{https://aclanthology.org/2024.eacl-long.13/}

\bibitem[{Zhang et~al.(2024)Zhang, Yilmaz, and Liu}]{zhang2025ilpxai}
Zhang Z, Yilmaz L, Liu B (2024) A critical review of inductive logic programming techniques for explainable ai. IEEE Transactions on Neural Networks and Learning Systems 35(8):10220--10236. \doi{10.1109/TNNLS.2023.3246980}

\bibitem[{Zheng et~al.(2023)Zheng, Chiang, Sheng, Zhuang, Wu, Zhuang, Lin, Li, Li, Xing et~al.}]{zheng2023judging}
Zheng L, Chiang WL, Sheng Y, et~al (2023) Judging llm-as-a-judge with mt-bench and chatbot arena. Advances in Neural Information Processing Systems 36:46595--46623

\bibitem[{Zuo et~al.(2024)Zuo, Velikanov, Rhaiem, Chahed, Belkada, Kunsch, and Hacid}]{zuo2024falcon}
Zuo J, Velikanov M, Rhaiem DE, et~al (2024) Falcon mamba: The first competitive attention-free 7b language model. \urlprefix\url{https://arxiv.org/abs/2410.05355}, {\href{https://arxiv.org/abs/2410.05355}{{arXiv:2410.05355}}}

\end{thebibliography}
%% if required, the content of .bbl file can be included here once bbl is generated
%%\input sn-article.bbl

\end{document}